\documentclass[a4paper,authoryear,sort,compress]{cas-dc}

\usepackage[caption=false,font=normalsize,labelfont=sf,textfont=sf]{subfig}

\usepackage{graphicx}
\usepackage{fancyhdr}
\usepackage{geometry}
\usepackage{hyperref}
\usepackage{tabularx}
\usepackage{booktabs}
\usepackage{multirow}
\usepackage{array}
\usepackage{amsfonts}
\usepackage[table]{xcolor}
\usepackage{makecell} % 新增 makecell 宏包
\usepackage{algorithm}
\usepackage{algpseudocode}
\usepackage{svg}
\usepackage{setspace}
\usepackage[authoryear]{natbib}

\usepackage{pifont}
\newcommand{\cmark}{\ding{51}} % ✓ 符号
\usepackage{amsmath,amsfonts}
\usepackage{array}
\usepackage{textcomp}
\usepackage{stfloats}
\usepackage{url}
\usepackage{verbatim}
\usepackage{float} 
\usepackage{multicol}
\usepackage{etoolbox}
\usepackage{xcolor}
\usepackage[table]{xcolor}

\begin{document}

\shorttitle{}
\shortauthors{Tang, Lu et al.}
\pagestyle{plain}
% \shorttitle{Dual-branch Maritime SORT (DMSORT): A Robust Multi-Object Tracking Framework for Maritime Scenes (Corrected Version)}
% \shortauthors{Tang, Lu et al.}

\title[mode = title]{DMSORT: An efficient parallel maritime multi-object tracking architecture for unmanned vessel platforms}
% \author{XiaoTong Gu\textsuperscript{1}
% \authorskip Shengyu Tang\textsuperscript{2}\thanks{Correspondence to:tangshengyu@ouc.edu.cn} 
% \authorskip Yiming Cao\textsuperscript{1}
% \authorskip Changdong Yu\textsuperscript{1} \\[0.5mm]
% {
% \fontsize{10.4pt}{9.84pt}\selectfont
% \textsuperscript{1}Dalian Maritime University \hspace{5.5mm} \textsuperscript{2}Ocean Univesity of China
% }\\
% }
\author[1]{Shengyu~Tang}\fnmark[1]
\author[2]{Zeyuan~Lu}\fnmark[1]
\author[3]{Jiazhi~Dong}
\author[4]{Changdong~Yu}[orcid=0000-0002-5759-4589]
\fnmark[*]
\ead{ycd@dlmu.edu.cn}
\author[1]{Xiaoyu~Wang}
\author[1]{Yaohui~Lyu}
\author[5]{Weihao~Xia}

% \fntext[equal1]{Shengyu Tang and Zeyuan Lu contributed equally to this work.}
% % \cortext[cor1]{Corresponding author}

\affiliation[1]{organization={College of Information Science and Engineering, Ocean University of China},
            city={Qingdao}, postcode={266100}, state={Shandong}, country={China}}
\affiliation[2]{organization={Marine Engineering College, Dalian Maritime University},
            city={Dalian}, postcode={116026}, state={Liaoning}, country={China}}
\affiliation[3]{organization={CSSC Haiwei Tech Co., Ltd},
            city={Zhengzhou}, postcode={450000}, state={Henan}, country={China}}
\affiliation[4]{organization={College of Artificial Intelligence, Dalian Maritime University},
            city={Dalian}, postcode={116026}, state={Liaoning}, country={China}}
\affiliation[5]{organization={College of Marine Electrical Engineering, Dalian Maritime University},
            city={Dalian}, postcode={116026}, state={Liaoning}, country={China}}

\begin{abstract}
Accurate perception of the marine environment through robust multi-object tracking (MOT) is essential for ensuring safe vessel navigation and effective maritime surveillance. However, the complicated maritime environment often causes camera motion and subsequent visual degradation, posing significant challenges to MOT. To address this challenge, we propose an efficient Dual-branch Maritime SORT (DMSORT) method for maritime MOT. The core of the framework is a parallel tracker with affine compensation, which incorporates an object detection and re-identification (ReID) branch, along with a dedicated branch for dynamic camera motion estimation. Specifically, a Reversible Columnar Detection Network (RCDN) is integrated into the detection module to leverage multi-level visual features for robust object detection. Furthermore, a lightweight Transformer-based appearance extractor (Li-TAE) is designed to capture global contextual information and generate robust appearance features. Another branch decouples platform-induced and target-intrinsic motion by constructing a projective transformation, applying platform-motion compensation within the Kalman filter, and thereby stabilizing true object trajectories. Finally, a clustering-optimized feature fusion module effectively combines motion and appearance cues to ensure identity consistency under noise, occlusion, and drift. Extensive evaluations on the Singapore Maritime Dataset demonstrate that DMSORT achieves state-of-the-art performance. Notably, DMSORT attains the fastest runtime among existing ReID-based MOT frameworks while maintaining high identity consistency and robustness to jitter and occlusion. 
\textbf{Code is available at:} \href{https://github.com/BiscuitsLzy/DMSORT-An-efficient-parallel-maritime-multi-object-tracking-architecture-}{\textcolor{red}{https://github.com/BiscuitsLzy/DMSORT-An-efficient-parallel-maritime-multi-object-tracking-architecture-}}.
\end{abstract}


\maketitle

\section{Introduction}\label{sec1}
As the exploitation of marine resources continues to grow, maritime security measures are constantly updated, driving an increasing demand for unmanned surface vessels (USVs) and marine robots to monitor and track the trajectories of ships, fishing vessels, and offshore platforms in predefined waters~\citep{1, 2, 3}. Maritime multi‐object tracking (MOT) plays a pivotal role in various applications such as intelligent surveillance, autonomous navigation, and maritime situational awareness~\citep{4,5,6,7,8}. While satellite remote sensing remains a primary tool for wide‐area surveillance, its effectiveness is often hindered by cloud cover and revisit limitations~\citep{9,10}. In contrast, vision‐equipped USVs can capture real‐time imagery at sea level, overcoming these constraints~\citep{11}. However, the unique challenges of maritime environments—unpredictable camera perspectives, low visibility, heavy occlusions, and drastic target scale variations—make MOT particularly difficult~\citep{12}. Unreliable maritime communications and poor signal quality further hamper real-time data transfer, rendering cloud-based processing impractical in many scenarios. Moreover, the computational constraints of many onboard systems complicate the deployment of resource-heavy models, highlighting the need for lightweight and efficient solutions~\citep{13,14,15}.

Recent approaches to MOT typically follow one of two paradigms: tracking-by-detection (TBD) or joint detection and tracking (JDT)~\citep{16}. JDT frameworks~\citep{17,18,19,20}  aim to unify detection and association within a single network, enabling end-to-end training. However, this approach often struggles with learning shared feature representations, particularly in maritime environments where small, low-resolution targets are prevalent. The simultaneous training of detection and appearance feature extraction networks can result in gradient interference, thereby limiting the quality of both object detection and appearance modeling. This conflict undermines the ability to effectively model object motion and appearance, resulting in inaccuracies in both.

In contrast, TBD methods~\citep{17,18,19} decouple the detection and tracking stages, employing robust detectors and modular tracking algorithms such as Kalman filtering. While this design offers better interpretability and flexibility, its sequential nature introduces latency. Moreover, in maritime environments, challenges such as motion blur, low resolution, and visual clutter significantly complicate the extraction of reliable motion and appearance features. These environmental difficulties hinder accurate object tracking and matching, making it challenging for TBD methods to recover correct identity associations. As a result, tracking performance is notably degraded, particularly under long-term occlusions or significant trajectory drift. Ultimately, both JDT and TBD methods face limitations in maritime settings.

From the above analysis, it becomes evident that the limitations of both JDT and TBD approaches arise from their incomplete use of motion and appearance cues. To further clarify this issue, we analyze the role of these two complementary sources of information in maritime MOT. In practice, motion models are typically built upon object detectors combined with Kalman filtering, where the detector provides current observations and the filter predicts target trajectories.While this design ensures short-term continuity, it is easily disrupted by ego-motion, unpredictable platform perturbations from wind and waves, and severe  scale variations of maritime targets, often leading to large prediction errors~\citep{21,22,23}. On the other hand, appearance cues are essential for long-term identity association, yet are often unreliable due to low resolution, motion blur, frequent occlusions, and cluttered backgrounds. Conventional CNN-based Re-ID models struggle with limited receptive fields~\citep{24},  while transformer-based approaches offer more robust representations~\citep{25,26,27,28} but remain computationally heavy under onboard resource constraints.

In addition to these challenges, the crux of maritime MOT lies in how motion and appearance information are fused. Traditional strategies often use simple weighted combinations of the two cues~\citep{20}, which fail under drift, noise, or long occlusions. This highlights the need for more adaptive fusion mechanisms that can dynamically balance unreliable cues, ensuring robust identity consistency in the face of occlusion, fragmentation, and adverse visual conditions.

Motivated by the above challenges, we propose a novel maritime multi-object tracking (MOT) framework that jointly enhances detection, tracking, and identity association under resource-constrained conditions. Specifically, we design a Dual-Branch Detection–Tracking Architecture (DDTA) that runs detection and tracking in parallel while isolating platform and target motion, enabling real-time perception without sacrificing accuracy.Building upon this architecture, we develop a Reversible Columnar Detection Network (RCDN), a compact multi-column detector with reversible feature refinement. It optimizes feature transmission to preserve information across layers, thereby achieving high recall and precision under extreme target scale variations while remaining lightweight. To address the limitations of conventional appearance modeling, we further introduce Li-TAE, a lightweight Transformer-based Re-ID module trained with adaptive hard triplets, which captures global contextual features and improves target discrimination under occlusion and low-resolution input. Finally, we propose a Clustering Optimized Feature Fusion (COFF) strategy that adaptively integrates motion and appearance cues using a nonlinear distance metric, unbiased momentum update, and spatial gating, thereby preserving identity consistency under noise, occlusion, and trajectory drift. Together, these components form a coherent solution that effectively tackles the unique difficulties of maritime MOT.

In summary, the main contributions of this work are as follows:
\begin{itemize}
    \item We design a Dual-Branch Detection–Tracking Architecture (DDTA) that decouples but parallelizes detection and tracking for real-time maritime perception.
    \item We develop a compact Reversible Columnar Detection Network (RCDN) that preserves shallow features for robust multi-scale detection under resource constraints.
    \item We introduce Li-TAE, a lightweight Transformer-based Re-ID module with adaptive hard triplet training for enhanced discrimination in occluded and low-resolution conditions.
    \item We propose a Clustering Optimized Feature Fusion (COFF) mechanism that adaptively balances motion and appearance cues to maintain identity consistency in challenging maritime environments.
\end{itemize}

\section{Related work}\label{sec2}
\subsection{Traditional Tracking Methods: From Correlation Filters to Deep Learning}

Before the advent of deep learning, research in visual tracking mainly focused on the single-object setting (SOT), as traditional detectors and hand-crafted features made large-scale multi-target tracking difficult. In this era, correlation filter-based methods such as MOSSE, CSK, and KCF~\citep{29,30,31,32} achieved high efficiency through Fourier domain regression and kernel learning, often running in real time. However, their reliance on shallow features and linear models limited robustness to occlusion, fast motion, and appearance variations, and they lacked mechanisms for identity management.

In contrast, early MOT approaches adopted a tracking-by-detection paradigm, where traditional detectors (e.g., HOG-SVM, DPM) provided candidate boxes, and association was performed across frames. Typical strategies included probabilistic models such as JPDA~\citep{33} and MHT~\citep{34}, global graph optimization via min-cost flow~\citep{35,36} and k-shortest paths~\citep{37}, and filtering-based methods using the Kalman filter~\citep{38} or particle filters~\citep{39}. While foundational, these approaches were hindered by the limited discriminative power of hand-crafted features, often resulting in ID switches and fragmented tracks.

The emergence of deep detectors and learned appearance embeddings transformed the field, shifting focus toward deep learning-based MOT. By combining accurate detection with robust identity features, these methods achieve far stronger performance, and can be broadly divided into motion-based trackers and hybrid motion–appearance trackers.

\subsection{Deep Learning-Based MOT: Motion and Appearance Fusion}

Motion-based MOT prioritizes efficiency, relying on object dynamics and inter-frame geometric consistency to associate detections. SORT~\citep{40} remains a widely used baseline, employing Kalman filtering and the Hungarian algorithm to achieve over 250 FPS. ByteTrack~\citep{17} improves robustness by incorporating low-confidence detections through a two-stage matching strategy. BOT-SORT~\citep{21} introduces optical flow-based background compensation and extends the Kalman state to better model dynamic camera settings. OCSORT~\cite{43} further refines motion modeling with observation center momentum (OCM) and online smoothing (OOS), enabling virtual motion estimation.

To overcome the limitations of motion-only trackers, hybrid methods fuse motion and appearance cues, typically using deep CNNs for appearance modeling. DeepSORT~\citep{20} enhances SORT with deep Re-ID embeddings for identity preservation. StrongSORT~\citep{18} introduces trajectory linking and Gaussian interpolation to improve association robustness. BoT-SORT-ReID~\citep{21} and ImprAssoc~\cite{44} incorporate dynamic feature weighting, while DeepOCSORT~\citep{45} applies adaptive cue balancing to dynamically adjust motion and appearance contributions.

Despite their improved robustness, most of these trackers adopt a modular, sequential pipeline, which limits real-time performance and struggles under extreme appearance degradation. Their reliance on convolutional appearance extractors also makes them vulnerable to occlusion, low resolution, and environmental interference.

\subsection{Domain-Specific Insights: Lessons from Benchmark Studies}

Benchmark-driven MOT research further illustrates the domain sensitivity of tracker performance. In DanceTrack~\citep{46}, where inter-target similarity and non-rigid motion dominate, traditional Re-ID pipelines degrade rapidly. SAM2MOT~\citep{47} tackles these challenges by employing segmentation-driven association and cross-object interaction mechanisms, effectively mitigating identity switches in such deformable and ambiguous motion settings. In the BEE24 benchmark, bee trajectories exhibit highly nonlinear dynamics, overwhelming classical filters. TOPICTrack~\citep{48} introduces attention-based appearance reconstruction and feature-level motion-appearance fusion to tackle biologically erratic motion.

These findings reveal two crucial limitations of existing MOT frameworks. First, most methods implicitly assume that targets follow predictable motion patterns with relatively stable appearances—assumptions that break down in scenes with non-rigid motion, crowded interactions, or occlusions. Second, classical association pipelines often lack the adaptability to handle dynamic appearance changes or severe occlusions, leading to fragmented trajectories and frequent identity switches.

These lessons translate directly to maritime environments, which introduce unique challenges: extreme scale variance (from large vessels to distant silhouettes), nonlinear motion due to wave action and platform drift, and frequent visual degradation from fog, sea-glare, and poor illumination. In such settings, objects rarely maintain consistent motion or appearance, and background dynamics further complicate association stability. As a result, conventional motion-based or hybrid trackers often struggle with ID fragmentation, tracking drift, and degraded robustness.

Furthermore, recent studies emphasize that tracking under non-rigid motion requires specialized modeling of both appearance variability and trajectory dynamics. Techniques such as attention-based reconstruction, multi-scale feature fusion, and trajectory smoothing have shown promise in biologically inspired benchmarks like BEE24 and DanceTrack. These insights suggest that maritime tracking systems could benefit from integrating similar strategies, such as adaptive feature learning, context-aware motion modeling, and hierarchical attention mechanisms, to better handle the compounded visual and kinematic distortions inherent to maritime surveillance.

\subsection{Maritime Multi-Object Tracking: Current Progress and Limitations}

To address these challenges, maritime-specific MOT frameworks have recently emerged. Sea-IoUTracker~\citep{3} introduces a buffered-IoU strategy and neighboring-frame recovery to stabilize associations under platform jitter. It also integrates a customized detector incorporating ASFF, CBAM, and YOLOv7, optimized for maritime targets. MOMT
~\citep{49} enhances ByteTrack by incorporating a fused-distance metric combining IoU, category, and object scale, alongside Gaussian cascade matching and an augmented Kalman filter with camera motion compensation. These strategies significantly improve maritime tracking robustness while maintaining real-time capability.

Nonetheless, existing maritime approaches remain grounded in terrestrial MOT frameworks, typically extending ByteTrack-like structures with heuristic enhancements. They often lack architectural flexibility and fail to jointly address scale-aware detection, appearance degradation, and dynamic fusion in an integrated fashion. Therefore, there is a growing need for end-to-end MOT systems that unify detection and tracking through parallelizable pipelines, adaptive multi-scale modeling, and transformer-based appearance extraction—capable of maintaining performance under the visual and kinematic distortions inherent to real-world maritime environments.

\section{Method}\label{sec3}
\begin{figure}
  \centering
  \includegraphics[width=0.9\linewidth]{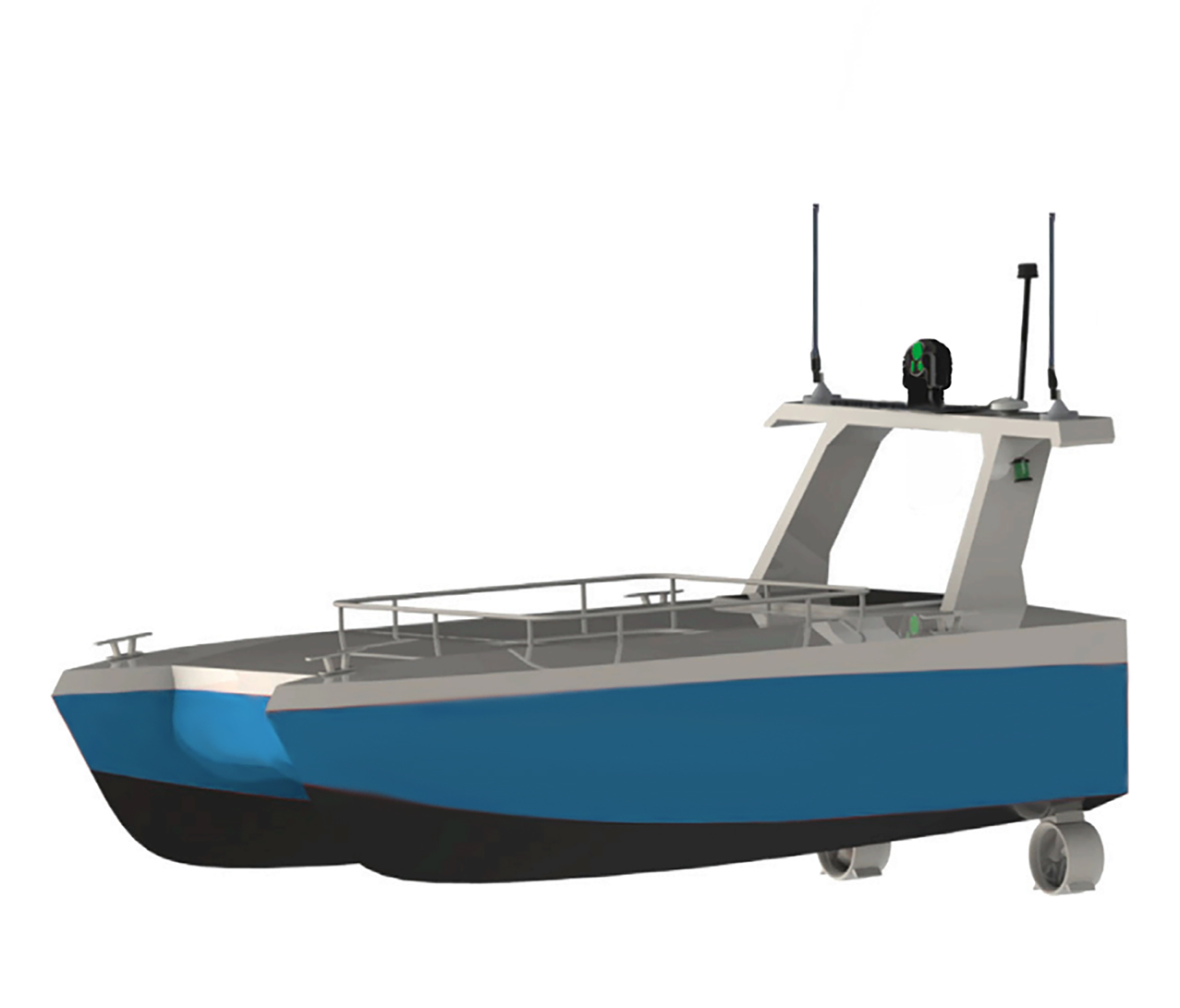}  
  \caption{Schematic diagram of unmanned vessel platform structure.}
  \label{fig:usv_demo}
\end{figure}

\begin{figure*}
  \centering 
  \includegraphics[width=0.95\textwidth]{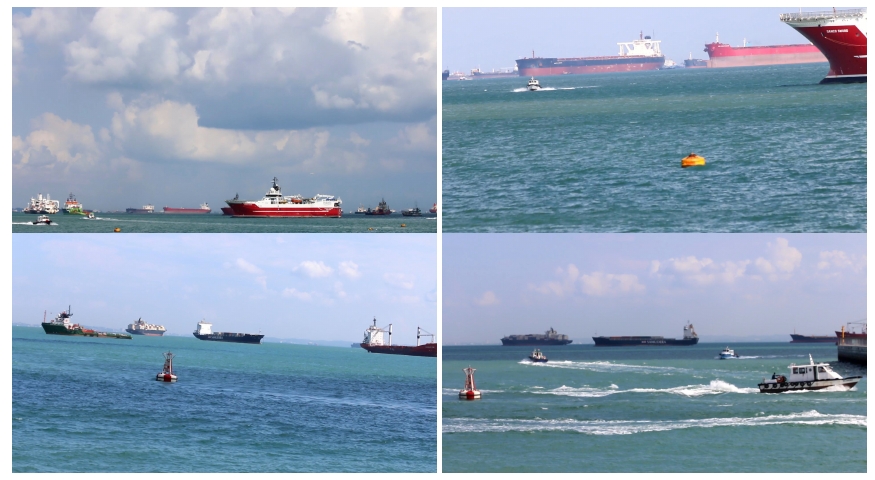}
  \caption{Illustrative examples of typical maritime scenes. These images showcase diverse oceanic scenarios, including vessels of various sizes, a floating buoy, and dynamic boat movements with wakes. Such scenes reflect the complex environments in maritime surveillance, posing challenges for object detection and tracking.}
  \label{fig:maritime_example}
\end{figure*}

Fig.~\ref{fig:usv_demo} presents a detailed schematic diagram of the structure and auxiliary devices of unmanned vessel platforms \citep{50,51}. The onboard cameras capture signals from the maritime environment and wirelessly transmit them to a host computer via a video surveillance transceiver. Once received, these video streams are processed by multi-object tracking (MOT) algorithms to identify and follow targets over time.

However, object detection and tracking in maritime environments present significant challenges. The motion of the platform introduces complex and irregular target movement patterns, while the large variation in object sizes—ranging from small buoys to large ships—further complicates reliable multi-object tracking (see Fig.~\ref{fig:maritime_example})~\citep{52}. To address these issues, we propose the DMSORT framework, which is specifically designed to enhance tracking performance under such challenging maritime conditions.

\subsection{Overview}
\begin{figure*}
    \centering
    \includegraphics[width=0.95\textwidth]{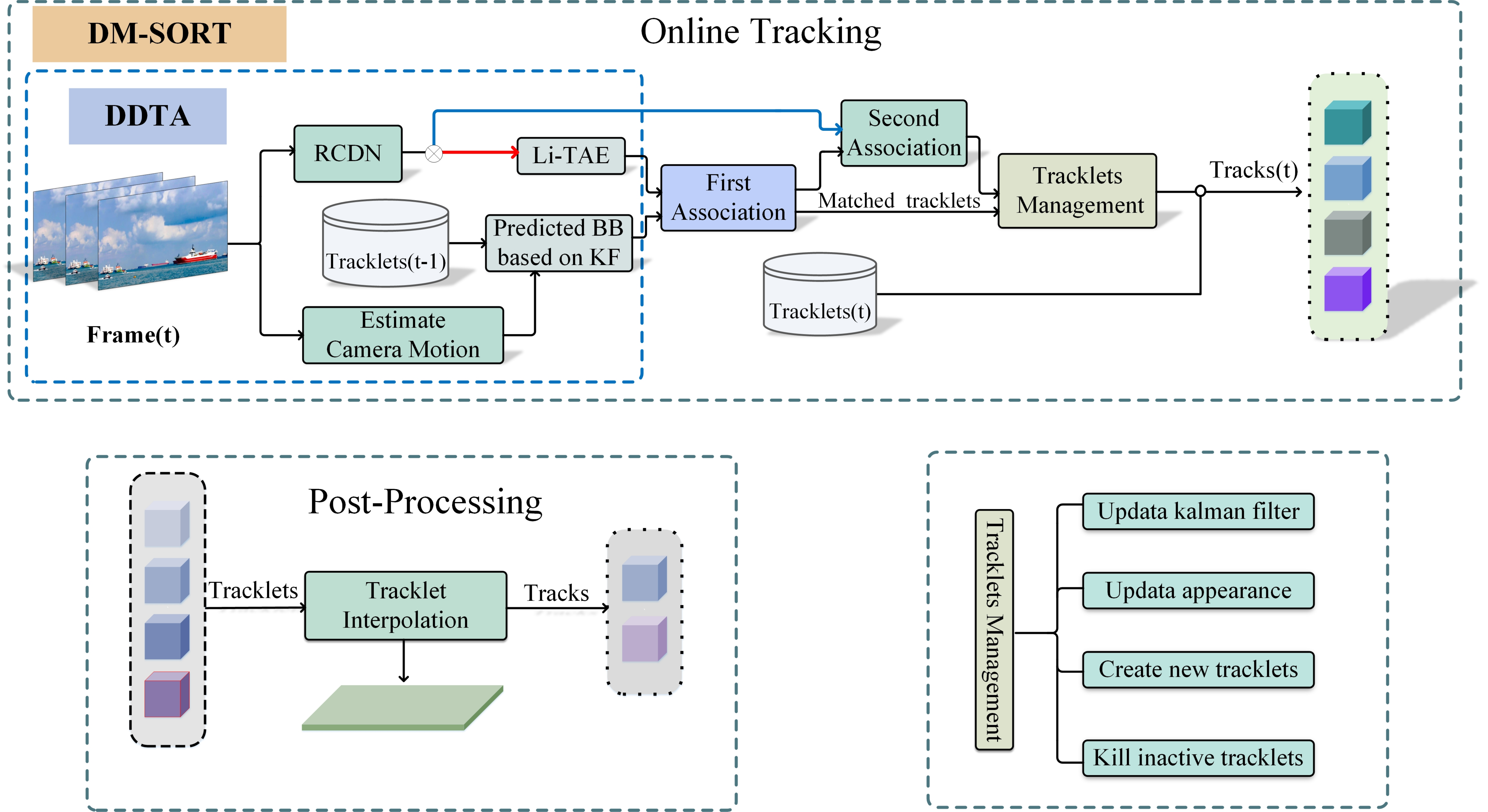}
    \caption{Overall pipeline of the proposed DMSORT framework. The system includes online tracking, where DDTA combines detection, motion compensation, and appearance encoding for two-stage association and tracklet management, and an optional post-processing step, where tracklet interpolation refines fragmented trajectories into final tracks.}
    \label{fig:overview}
\end{figure*}

The overall pipeline of the DMSORT framework is shown in Fig.~\ref{fig:overview}, consisting of online tracking and post-processing. The online module adopts Detection-driven Dual-Task Association (DDTA), which integrates motion compensation and appearance feature extraction for robust tracking.

In DDTA, an RCDN-based detector generates bounding boxes with different confidence levels. For high-confidence detections, a lightweight Transformer encoder (Li-TAE) extracts discriminative appearance features for reliable association.Meanwhile, the observed motion in the video results from the superposition of the platform’s movement and each target’s intrinsic motion. To decouple these components, the motion branch computes the affine transformation between successive frames~\citep{53,54,55} to estimate the platform’s motion. Once obtained, the affine matrix is incorporated into the Kalman filter’s prediction step by multiplying it with the predicted state vector and covariance matrix. This correction compensates for camera-induced motion, aligns target trajectories with a stabilized reference frame, and enables more accurate modeling of each target’s true trajectory under platform-driven disturbances.

After feature extraction and motion correction, a two-stage data association is performed. The COFF scheme first matches high-confidence detections to predicted tracks by fusing Re-ID features and IoU via the Hungarian algorithm. Remaining tracks are then associated with low-confidence detections using IoU. Finally, the trajectory management module updates motion states and features, initializes new tracks, and removes inactive ones. Overall, DMSORT achieves reliable multi-object tracking under camera jitter, occlusion, and appearance degradation.

\subsection{Reversible Columnar Object Detection Network}

\begin{figure*}
    \raggedright
    \includegraphics[width=\textwidth]{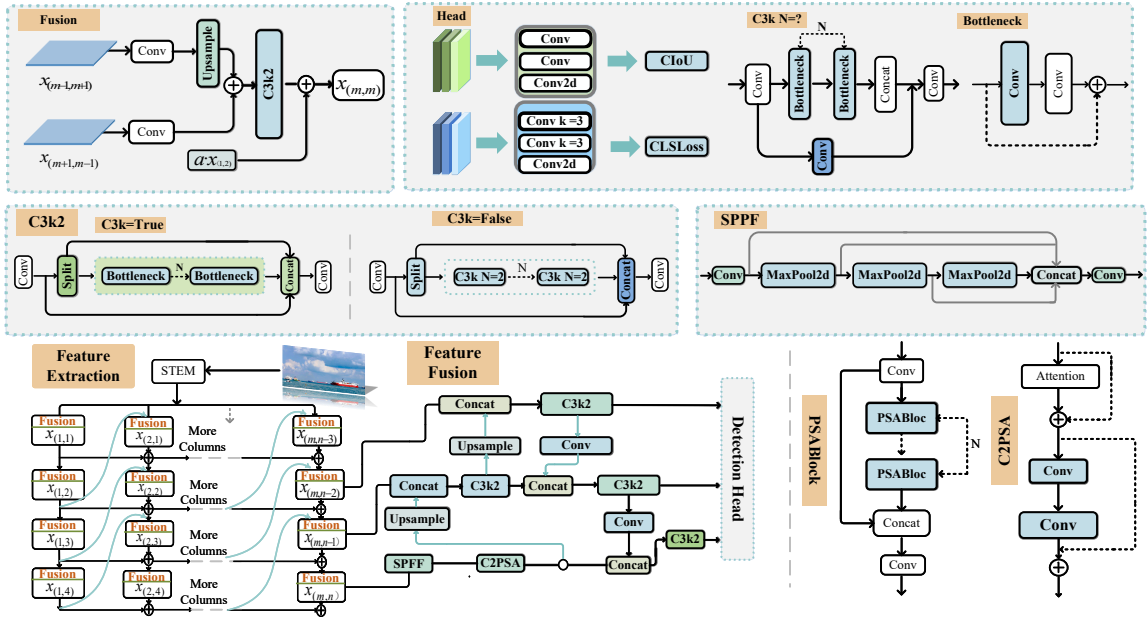}
    \caption{Architecture of the proposed RCDN. The left-bottom part illustrates the reversible column backbone, where each level exchanges features across scales for lossless extraction. The middle modules (SPPF, C3k2, C2PSA) perform multi-scale fusion and attention refinement. The right detection head outputs classification and regression for final object prediction.}
    \label{fig:wideimage}
\end{figure*}

This study proposes a multi-level feature decoupling object detection framework  Reversible Column Detection Network (RCDN). The core innovation of RCDN lies in its lossless feature transformation, which is guided by residual connections and a reversible structure that provides a mathematical guarantee for explicitly reversible approximations.

As illustrated in Fig.\ref{fig:wideimage}, the network is divided into three major components: the Hierarchical Feature Extraction, Feature Fusion, and Detection Head, which work together to process input data efficiently.

\begin{algorithm}
\caption{Hierarchical Feature Extraction}

\normalsize 
 \begin{spacing}{1} 
\label{alg:revcol}
\begin{algorithmic}[1]
\State Conv\textsubscript{stem}: Conv(k=4, s=4, p=0)
\State Block\textsubscript{level}: C3K2
\State $\alpha_{level}$: learnable scalar

\Function{Fusion}{$down, up, first\_col$}
    \State $down\_branch \gets \text{Conv(k=2, s=2, p=0)}(down)$
    \If{$first\_col$}
        \State \Return $down\_branch$
    \Else
        \State $up\_branch \gets \text{Upsample}(\text{Conv(k=3, s=1, p=1)}(up))$
        \State \Return $down\_branch + up\_branch$
    \EndIf
\EndFunction

\Function{RevColForward}{$x$}
    \State $x \gets \text{Conv}_{stem}(x)$ \Comment{Patch embedding}
    \State Initialize $c_0, c_1, c_2, c_3 \gets 0$
    \For{$i = 1$ to $N$}
        \State $first\_col \gets (i == 1)$
        \For{$level = 0$ to $3$}
            \If{$level == 0$}
                \State $down \gets x$
            \Else
                \State $down \gets c_{level - 1}$
            \EndIf

            \If{$level < 3$ \textbf{and} $\neg first\_col$}
                \State $up \gets c_{level + 1}$
            \Else
                \State $up \gets \text{None}$
            \EndIf

            \State $fused \gets \text{Fusion}(down, up, first\_col)$
            \State $out \gets \text{Block}_{level}(fused)$
            \State $c_{level} \gets \alpha_{level} \cdot c_{level} + out$
        \EndFor
    \EndFor
    \State \Return $[c_0, c_1, c_2, c_3]$
\EndFunction
\end{algorithmic}
  \end{spacing}
\end{algorithm}

The first component of the proposed network, termed Hierarchical Feature Extraction, is detailed in Algorithm~\ref{alg:revcol}. In summary, this component consists of $N$ weight-heterogeneous sub-networks that progressively extract multi-scale features through a hierarchical structure. Each sub-network, referred to as a "column", contains four stages ($C_1$ to $C_4$), which capture spatial semantics at increasingly higher levels of abstraction. These stages are designed to progressively refine the feature representation as the data passes through, using a residual coupling mechanism that fuses intra- and inter-column information:

{
\small
\begin{equation}
x_{(m, n)} = \alpha \cdot x_{(m-1, n)} + \phi \left( g \left( x_{(m, n-1)} \right) \oplus h \left( x_{(m-1, n+1)} \right) \right)
\end{equation}
}

Here, $g(\cdot)$ and $h(\cdot)$ denote $2 \times$ downsampling and $2 \times$ upsampling operators, respectively; $\phi(\cdot)$ represents a non-linear transformation using the $C_3 K_2$ module, and $\alpha$ is a learnable scaling factor regulating residual propagation.

This approach facilitates intra-scale residual transmission via $\alpha \cdot x_{(m-1, n)}$, preserving feature continuity and minimizing semantic degradation as the network progresses through its layers. Simultaneously, cross-scale feature fusion is achieved through bidirectional sampling between neighboring columns, $(x_{(m-1,n+1)},$ $x_{(m,n-1)})$, which enhances multi-level feature representation.

To ensure lossless feature recovery and enable the reconstruction of previous features during reverse inference, we design the update process as an invertible function. This formulation allows the reconstruction of the earlier features from the final column output using:
{
\footnotesize
\begin{equation}
x_{(m-1, n)} = \left[ x_{(m, n)} - \phi \left( g \left( x_{(m, n-1)} \right) \oplus h \left( x_{(m-1, n+1)} \right) \right) \right] / \alpha
\end{equation}
\label{anti_trans}
}

The proposed architecture, incorporates a boundary condition ($h(x) = 0$ when $n = n_{\max}$ and $m = 1$) to ensure the uniqueness of the feature reconstruction process. This formulation enables the explicit recovery of earlier features from final-layer representations through the invertible transformation defined in Eq.~\ref{anti_trans}. A key practical advantage of this property is that it eliminates the need to store intermediate activations during the backward pass—features can be recomputed on-the-fly via the inverse operation, leading to substantially reduced memory consumption during training~\citep{23}.

Furthermore, the universal approximation theorem ensures that the network can learn to approximate inverse mappings autonomously. Formally, the learned transformation $\boldsymbol{T}_{\boldsymbol{\theta}}$ satisfies~\citep{56,57}.

\begin{equation}
\left|\boldsymbol{T}{\boldsymbol{\theta}}^{-1}\left(x{(M, n)}\right) - x_{(m-1, n)}\right| < \epsilon
\end{equation}

for any small tolerance $\epsilon > 0$, where $x_{(M, n)}$ denotes the output feature and $x_{(m-1, n)}$ the corresponding earlier feature. Thus, even in the absence of an explicitly programmed inversion, the model can discover—through training—efficient strategies to reconstruct or leverage multi-level features directly from the output, potentially surpassing the performance achievable through the hand-designed inverse alone.

To enable accurate inverse transformation fitting during inference, we introduce the Feature Fusion module. This module integrates multi-scale features from the network's final stages, ensuring feature consistency across resolutions and facilitating reverse mapping during inference.

In the Feature Fusion process, multi-scale features are produced at different resolutions, denoted as ${x_{(M, 1)}, x_{(M, 2)}, x_{(M, 3)}, x_{(M, 4)}}$, with downsampling ratios ranging from $1/8$ to $1/64$. At the final output feature $x_{(M, 4)}$, Spatial Pyramid Pooling Fast (SPPF) is applied. The SPPF layer uses pooling kernels of sizes $5 \times 5$, $9 \times 9$, and $13 \times 13$ to capture global contextual information while preserving important fine-grained details across multiple scales.

Following the application of SPPF, the output feature map undergoes attention weighting through the C2PSA module. The attention mechanism helps enhance important features by applying self-attention and feed-forward operations. The C2PSA module first splits the feature map into two parts and processes one part with a series of PSA blocks for attention and transformation. The processed feature is then recombined and passed through a final convolution layer to produce the refined features.

After feature refinement, the aggregated features are fused across scales using a bidirectional feature fusion process. This is achieved through a modified PANet architecture. In the top-down pathway, higher-level features are upsampled and fused with shallower representations, while in the bottom-up pathway, lower-level features are downsampled and combined with deeper layers to enhance semantic consistency across scales.

These features are then fed into a YOLOv11's detection head, where the classification and regression tasks are optimized independently. This design maintains real-time inference capability while improving detection accuracy, particularly in dense and small-object scenes.

While the initial feature extraction provides a performance floor through lossless transmission, the Feature Fusion module surpasses this bound by enhancing multi-scale aggregation and non-linear expressiveness. This improves the model’s ability to capture complex patterns, achieving more accurate detection, especially in challenging object-dense environments.

\subsection{Lightweight Transformer-Based Appearance Extractor}

\begin{figure*}
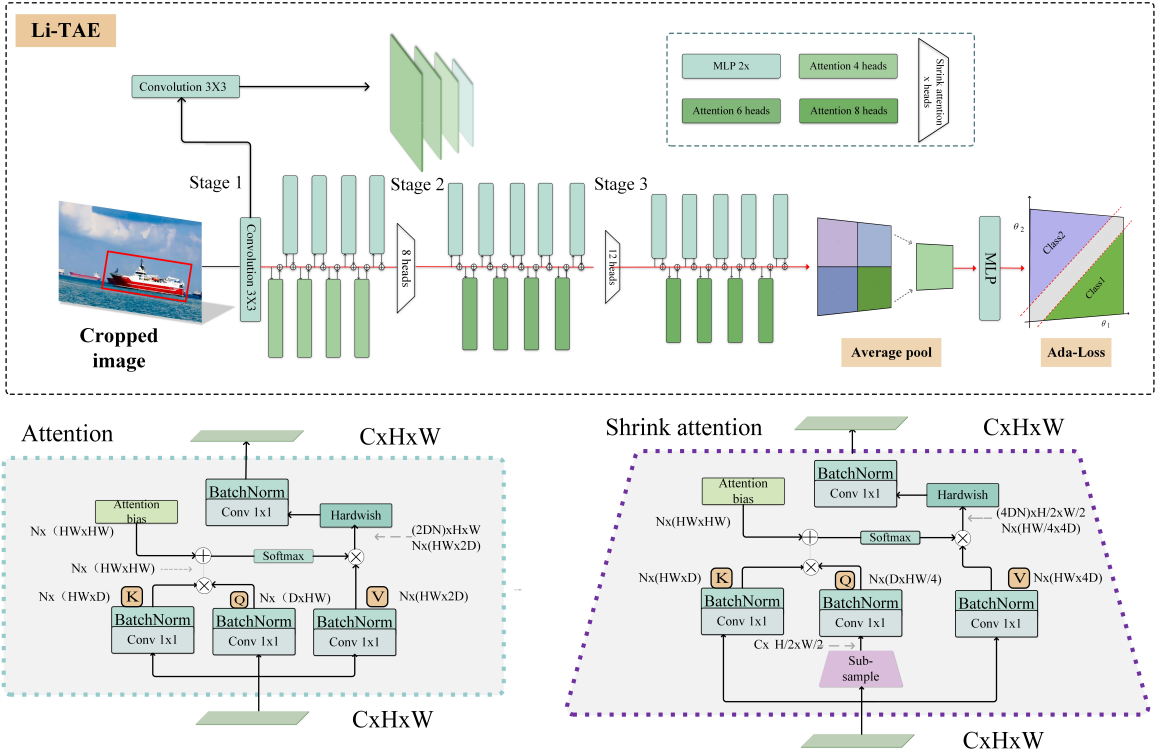

  \centering
  \includegraphics[width=0.95\textwidth]{zhuti.pdf}\\[4pt]
  \includegraphics[width=0.95\textwidth]{zuizhong.pdf}
  \caption{Architecture of the proposed Li-TAE module. The figure illustrates the overall pipeline and detailed structure of the lightweight Transformer-based appearance encoder, including the use of self-attention with our designed positional embeddings and shrink attention.}
  \label{fig:li-tae}
\end{figure*}

To extract discriminative appearance features for maritime multi-object tracking—characterized by low resolution, occlusion, and background clutter—we propose the Lightweight Transformer-based Appearance Extractor (Li-TAE). As shown in Fig.~\ref{fig:li-tae}, this module leverages a Vision Transformer (ViT) architecture enhanced with self-attention mechanisms for efficient feature extraction. It overcomes the limitations of CNN-based ReID backbones, which often struggle with limited receptive fields and insufficient global context modeling.

The process begins with an image embedding module that downsamples the input image (224×224×3) using  four layers of 3×3 convolutions with stride 2, producing a compact 14×14×256 feature map. These embeddings are then processed by customized Transformer blocks, progressively refining the visual features.

We introduce key architectural modifications to the ViT model. First, the classification head is replaced with a task-specific multi-layer perceptron (MLP) to produce compact identity-preserving embeddings instead of classification logits. Second, the Transformer blocks adopt multi-head self-attention with learnable relative positional biases, allowing translation equivariance while capturing geometric relationships. Specifically, the attention score between spatial positions $(x, y)$ and $(x', y')$ for attention head $h$ is computed as:

\begin{equation}
A_{(x,y),(x',y')}^h = Q_{(x,y),:} \cdot K_{(x',y'),:} + B_{|x-x'|,|y-y'|}^h
\end{equation}

This replaces fixed positional encodings with learnable biases conditioned on spatial displacement, improving generalization under viewpoint shifts and target displacements.

As shown in the fig.\ref{fig:li-tae}, to further enhance efficiency , we adopt shrinking attention blocks that downsample the feature maps by applying strided operations on the queries (Q). This reduces the feature map size, with the output transforming from an input tensor of size (C, H, W) to (2C, H/2, W/2). Lightweight MLP layers—implemented as 1×1 convolutions followed by batch normalization—replace dense layers, with the expansion ratio reduced from 4 to 2, optimizing the trade-off between parameter size and representational power.

The final Transformer output undergoes global average pooling, followed by a linear projection via the output MLP, resulting in a 512-dimensional embedding vector. This vector is designed to serve as a compact descriptor for identity discrimination in the appearance matching pipeline.

In terms of training objectives, we propose an adaptive angular margin loss (Ada-Loss) to replace conventional triplet loss, which is prone to overfitting and gradient instability in deep Transformer models. This loss operates on the hypersphere of L2-normalized features, enforcing angular separation between positive and negative pairs:

\begin{align}
\theta_1 &= \arccos(\vec{a} \cdot \vec{p}) + \theta, \theta_2 = \arccos(\vec{a} \cdot \vec{n}) \\
d_i &= 1 - \cos\theta_i\quad (i=1,2) \\
\text{loss} &= \left[ \max(d_1 - d_2 + \alpha, 0) \right]^2
\end{align}

Where $\theta_1$ and $\theta_2$ are the angles between the query vector $\vec{a}$ and the positive vector $\vec{p}$, and the negative vector $\vec{n}$, respectively. The margin $\alpha$ controls the angular separation between positive and negative pairs.

The gradient of the loss function, as shown in the following equation:

$$
\nabla L=2\left(d_1-d_2+\alpha\right)\left(\nabla d_1-\nabla d_2\right)
$$

This demonstrates the benefits of dynamic hard sample mining. The gradient magnitude increases with the severity of margin violations, ensuring that hard examples receive larger gradients, which accelerates convergence. Additionally, the loss incorporates adaptive gradient decay, reducing updates for well-separated samples and addressing potential convergence issues in later training stages. This mechanism can be viewed as actively resampling difficult examples, enhancing the model's robustness and learning efficiency.

Furthermore, the angular formulation aligns with the L2-normalized output embeddings, and the margin parameter $\alpha$ allows fine-grained control over the compactness and discrimination of the learned feature space.

\subsection{Clustering Optimized Feature Fusion}

To address challenges in multi-object tracking (MOT) within maritime environments, we introduce a clustering optimized feature fusion (COFF) framework. This approach enhances robustness against occlusion, appearance ambiguity, and environmental degradation. Our method incorporates temporal smoothing, nonlinear feature scaling, and spatial gating, ensuring more reliable identity association under complex conditions.

In maritime scenarios, the relatively stable environment leads to low inter-frame motion, enabling the use of long-term appearance features to improve association reliability, even under partial occlusion or adverse weather. To exploit this continuity, we propose an Unbiased Exponential Moving Average (UEMA) for updating appearance representations of tracked targets.The UEMA applies exponential decay to historical features and incorporates a normalization factor to correct bias in early frames.Specifically, given the raw appearance feature $f_i^k$ of target $i$ at frame $k$, we compute the smoothed embedding $e_i^k$ and its unbiased estimate $\widehat{e_i^k}$ as follows:

\begin{align}
e_i^k &= \alpha e_i^{k-1} + (1 - \alpha) f_i^k \\
\widehat{e_i^k} &= \frac{e_i^k}{1 - \alpha^k}
\end{align}

Here, $\alpha$ denotes the temporal decay factor and the initial appearance embedding $e_i^0$ is set to zero. This formulation improves the temporal stability of appearance features without overly relying on noisy early observations, thus preserving identity continuity through prolonged occlusions or partial visual degradation.

Although temporal fusion improves feature consistency, we observe that cosine similarity, commonly used for appearance matching, often lacks sufficient discriminative power. Empirical analysis reveals that intra-class cosine distances average around $ 10^{-4}$, while inter-class distances cluster around $ 10^{-2}$. While this reflects a substantial relative difference, the small absolute gap limits the effectiveness of joint optimization with motion cues, which typically have distances in the range of $10^{-1}$. This imbalance leads to the dominance of appearance similarity terms, while the influence of motion information is suppressed in the overall association cost.

\begin{figure*}
  \centering
  \includegraphics[width=0.85\linewidth]{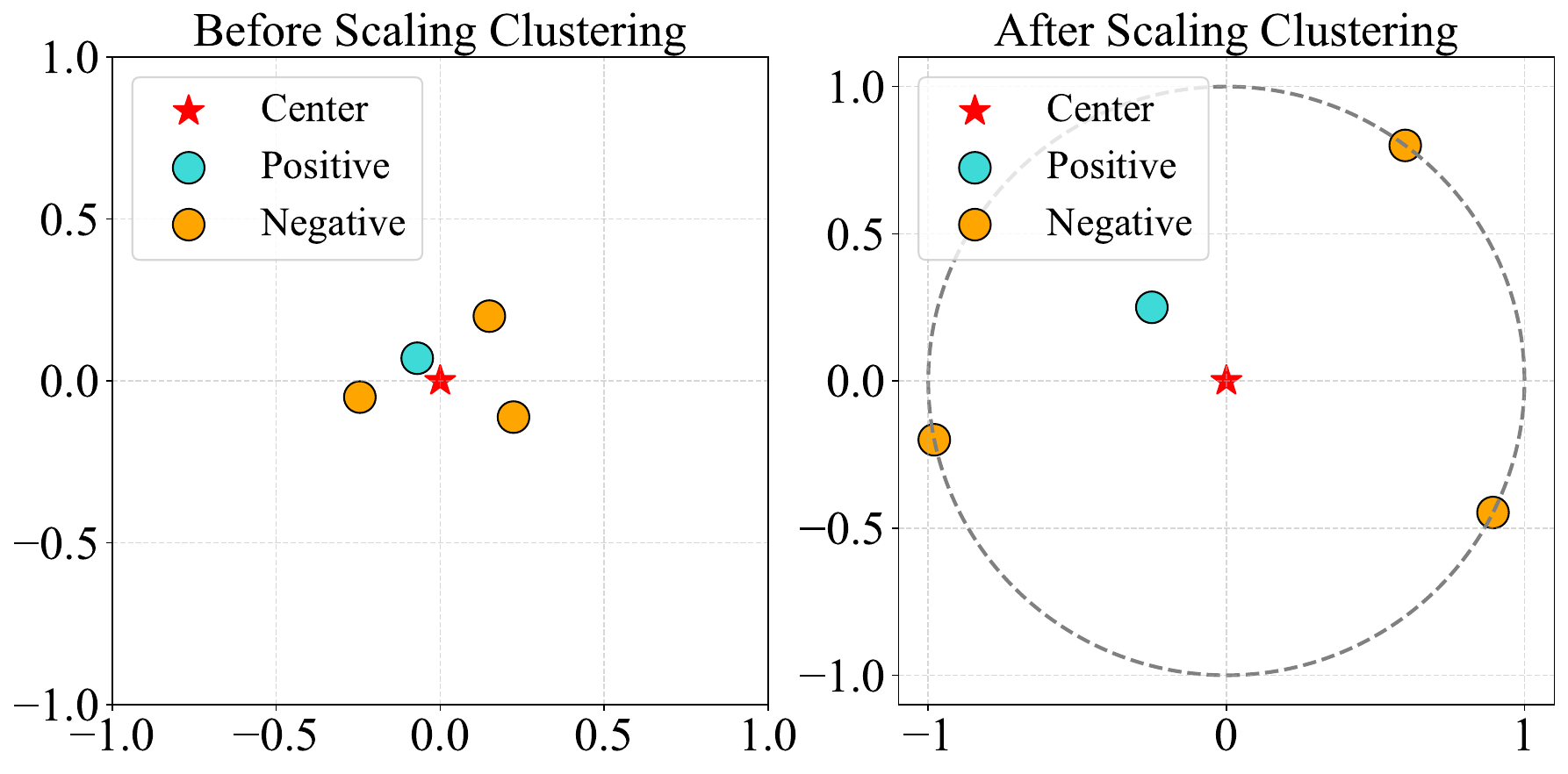}  
  \caption{Impact of scaling on distance metrics in multi-object tracking. The left panel displays samples before scaling, where both positive (cyan) and negative (orange) samples are near the center (red star). After scaling (right panel), the distance between samples is magnified, clearly distinguishing positive from negative samples, thus enhancing tracking accuracy.}
  \label{fig:zhiguanjvlei}
\end{figure*}

To address this issue, we propose a nonlinear amplification mechanism to rescale the cosine distance, improving its compatibility with motion-based metrics. Specifically, we multiply the original appearance distance by a scaling factor $\beta = 10^3$ and clip the result to a maximum of 1:

\begin{equation}
d_{i,j}^{\cos} = \min(d_{i,j}^{\cos} \cdot \beta, 1)
\end{equation}

As illustrated in Fig.~\ref{fig:zhiguanjvlei}, this transformation amplifies intra-class appearance distances to approximately $10^{-1}$, while inter-class distances move toward 1. After scaling, the similarity space becomes more separable, with positive pairs constrained within $[0, 0.2]$ and negative pairs concentrated near $[1 - \epsilon, 1]$, where $\epsilon$ is a small constant. This enhanced contrast significantly improves the ranking reliability of appearance-based associations.

To further suppress false positives caused by abrupt appearance changes or unreliable features, we introduce a spatial gating mechanism based on IoU. When the overlap between a track and a detection falls below a predefined threshold $\theta_{\mathrm{iou}} = 0.3$, their appearance similarity is overridden to the maximum dissimilarity value:

\begin{equation}
d_{i,j}^{\cos} = 1, \quad \text{if } d_{i,j}^{\text{iou}} < \theta_{\text{iou}}
\end{equation}

This gating strategy effectively filters out visually similar but spatially implausible matches, enhancing the robustness of the association process under occlusion or scene clutter.

Finally, the overall matching cost combines motion and appearance cues through a multiplicative formulation. For each pair of track $i$ and detection $j$, the final cost $C_{i,j}$ is computed as:

\begin{equation}
C_{i,j} = d_{i,j}^{\cos} \cdot d_{i,j}^{\text{iou}}
\end{equation}

This formulation ensures that both appearance dissimilarity and motion misalignment contribute to the association cost, making the system sensitive to both modalities. The scaling factor $\beta$ enhances separability without disrupting the relative order of association costs. Empirical results confirm that across a range of $\beta$ values (e.g., $\beta \in[200,2000]$ ), intra-class costs remain low, while inter-class costs align with motion-based distances, leading to stable matching performance across diverse tracking conditions.

In summary, by combining UEMA-based temporal smoothing, nonlinear distance scaling, and spatial gating, our fusion strategy significantly enhances identity association in maritime MOT, particularly under occlusions, noise, and scale imbalances between motion and appearance cues.

\section{Experiments}\label{sec4}
\subsection{Experimental Setup}

We evaluate our method on the Singapore Maritime Dataset~\citep{52}, a benchmark for maritime multi-object tracking (MOT). This dataset includes video sequences captured from static port surveillance cameras and mobile unmanned surface vehicles (USVs), featuring various environmental conditions (day, night, clear weather) at a resolution of $1920 \times 1080$ pixels. It introduces domain-specific challenges such as water reflections, dense vessel traffic, occlusions, and low-contrast targets, with annotations for seven maritime object categories: \textit{ferry}, \textit{buoy}, \textit{ship}, \textit{speedboat}, \textit{small boat}, \textit{kayak}, and \textit{sailboat}.

\begin{figure*}
  \centering
  \includegraphics[width=0.95\textwidth]{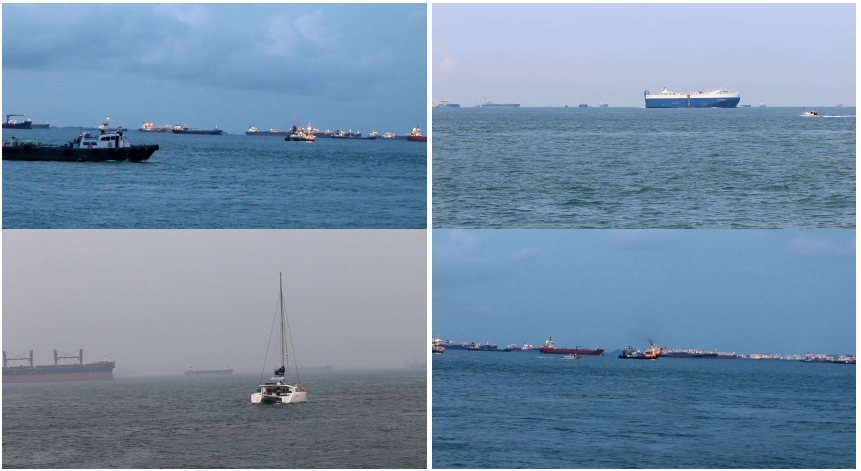}
  \caption{Four representative scenes from the Singapore Maritime Dataset illustrating the diversity of maritime conditions. The dataset encompasses a spectrum of scenarios, from night operations and foggy weather to open waters and crowded ports, capturing the complexity of real-world maritime environments.}
  \label{fig:maritime_challenges}
\end{figure*}

Fig.~\ref{fig:maritime_challenges} highlights the unique challenges of maritime environments, including densely packed small targets, motion blur, and severe occlusion. For object detection, we randomly sample temporally independent frames and divide them into training (60\%), validation (20\%), and testing (20\%) subsets, ensuring that temporally adjacent frames are not split across subsets. For Re-Identification (ReID), the first 50\% of frames in each video are used for training, while the remaining 50\% are reserved for validation.

To comprehensively evaluate tracking performance on the Singapore Maritime Dataset~\citep{58,59,60}, we adopt the CLEAR MOT evaluation framework~\citep{66}, which encompasses a suite of widely adopted tracking metrics. These include Multiple Object Tracking Accuracy (MOTA), which penalizes false positives (FP), false negatives (FN), and identity switches (IDs); IDF1~\citep{61}, which measures identity preservation accuracy; and Higher Order Tracking Accuracy (HOTA)~\citep{62}, which jointly evaluates detection, association, and localization quality. Additionally, we report Fragmentations (Frag), Localization Accuracy (LocA), Identity Switches (IDs), and runtime efficiency (FPS) to provide a holistic view of tracker performance.

To further validate the generalization ability of our model—especially under dense and non-maritime conditions—we also evaluate it on the widely adopted MOT17 benchmark~\citep{63}, which has become a de facto standard in the MOT community. MOT17 features dense pedestrian crowds with frequent occlusions and complex motion patterns, offering a challenging yet complementary testbed to maritime scenarios.

For fair comparison on MOT17, we follow the community-standard protocol by using the provided public Faster R-CNN detections. Moreover, in alignment with BOXMOT our chosen baseline, which reports only MOTA, IDF1, and HOTA—we restrict our evaluation on MOT17 to these three core metrics. This streamlined approach ensures a consistent and equitable comparison across methods.

The tracking metrics are computed using the following formulations:
\begin{equation}
\mathrm{HOTA} = \sqrt{\mathrm{DetA} \cdot \mathrm{AssA}} = \sqrt{\frac{\sum_{c \in \mathrm{TP}} A(c)}{\mathrm{TP} + \mathrm{FN} + \mathrm{FP}}}
\end{equation}
\begin{equation}
A(c) = \frac{\mathrm{TPA}(c)}{\mathrm{TPA}(c) + \mathrm{FPA}(c) + \mathrm{FNA}(c)}
\end{equation}
\begin{equation}
\mathrm{DetA} = \frac{|\mathrm{TP}|}{|\mathrm{TP}| + |\mathrm{FN}| + |\mathrm{FP}|}
\end{equation}
\begin{equation}
\mathrm{MOTA} = 1 - \frac{\sum_{t} (\mathrm{FP}_t + \mathrm{FN}_t + \mathrm{IDS}_t)}{\sum_{t} G_t}
\end{equation}
\begin{equation}
\mathrm{IDF1} = \frac{2 \cdot \mathrm{IDTP}}{2 \cdot \mathrm{IDTP} + \mathrm{IDFP} + \mathrm{IDFN}}
\end{equation}
\begin{equation}
\mathrm{LocA} = \frac{1}{N} \sum_{i=1}^{N} \mathrm{IoU}(GT_i, Det_i)
\end{equation}

Here, $N$ denotes the number of true positive detections, and $\mathrm{IoU}(GT_i, Det_i)$ represents the intersection-over-union between the predicted and the ground-truth bounding boxes.

In summary, higher values of HOTA, MOTA, IDF1, and LocA indicate better tracking performance in terms of detection quality, identity association, and spatial localization. Conversely, lower values of FP, FN, Frag, and IDs correspond to more accurate and stable tracking. FPS serves as an indicator of system efficiency and real-time capability.

\subsection{Implementation Details}

All experiments are conducted on a workstation equipped with an NVIDIA GeForce RTX 4090D GPU, which provides a consistent and high-performance environment for model training and evaluation. To simulate deployment in resource-constrained maritime scenarios, the inference speed (FPS) in model comparison experiments is additionally measured on a laptop equipped with an NVIDIA GeForce RTX 4060 GPU. This setup demonstrates the real-time capability and efficiency of our method under limited computational resources.

The object detector is built upon the proposed Reversible Columnar Detection Network (RCDN), which enhances multi-scale feature fusion for robust maritime target detection. Feature maps extracted by RCDN are forwarded to the native decoupled classification head of YOLOv11 for object categorization. For appearance modeling, the Li-TAE module is adopted as the ReID backbone, designed to generate compact and discriminative identity embeddings.

The detector is trained for 300 epochs using the AdamW optimizer with a momentum of 0.937 and a batch size of 32. The initial learning rate is set to 0.01 and gradually reduced to 0.0001 following a cosine annealing schedule. A weight decay of 0.0005 is applied for L2 regularization, and a warm-up phase of 3 epochs is used at the beginning of training. The loss function adopts a weighted scheme, where the box regression loss is weighted at 7.5 and the classification loss at 0.5. To improve robustness against occlusion and scale variation, data augmentation techniques including mosaic augmentation (probability = 1.0) and random rotation within $\pm10^{\circ}$ are applied.

The ReID feature extractor is trained separately for 300 epochs using a hard triplet mining strategy to enhance inter-class separability and intra-class compactness. Input images are resized to $224 \times 224$ using bilinear interpolation and augmented with the \texttt{rand-m9-mstd0.5} policy. Training employs an angular margin loss with a scaling factor $s=64$ and a margin $m=0.5$. The batch size is set to 64, and the learning rate is fixed at $10^{-3}$, with a 30-epoch warm-up followed by cosine annealing.

During inference, the detection confidence threshold is set to 0.6. For data association, a dual-constraint matching scheme is employed. The linear matching stage uses an appearance similarity threshold of 0.25 and an IoU threshold of 0.75. A two-stage matching strategy is implemented with primary and secondary thresholds set to 0.5 and 0.1, respectively. The appearance distance is scaled by a factor of 800, and the final matching score threshold is 0.85. To handle short-term occlusions, a trajectory buffer of 30 frames is maintained. No post-processing is applied during evaluation to reflect real-world deployment conditions.

\subsection{ Comparison Experiment}
\begin{table*}
  \centering
  \caption{MOT results on the Singapore Maritime Dataset. * indicates methods enhanced with RCDN and Li-TAE where Re-ID is available; RCDN only otherwise. Best scores before and after our improvements are bolded.}
  \label{tab:singapore_maritime}
  \setlength{\tabcolsep}{4pt}
  \renewcommand{\arraystretch}{1.2}
  \resizebox{0.9\textwidth}{!}{%
  \begin{tabular}{@{}l|c|cccccccccc@{}}
    \hline
    Method & Re-ID & HOTA$\uparrow$ & LocA$\uparrow$ & FN$\downarrow$ & FP$\downarrow$ & Frag$\downarrow$ & IDF1$\uparrow$ & IDs$\downarrow$ & MOTA$\uparrow$ & FPS$\uparrow$ \\
    \hline
    % ---------------- 不带星号 ----------------
     BOT-SORT~\cite{21}      & \cmark & 72.13 & 92.90 & 19611 & 8783 & \textbf{25} & 78.81 & \textbf{162} & 70.62 & 20.4 \\
    OCSORT~\citep{43}        & --     & 71.18 & 92.54 & 19980 & \textbf{7421} & 135 & 78.43 & 180 & 70.62 & \textbf{47.4} \\
    DeepOCSORT~\citep{45}    & \cmark & 72.48 & 91.61 & 17035 & 8323 & 114 & 80.44 & 186 & 73.74 & 25.9 \\
    ImprAssoc~\citep{44}     & \cmark & 71.56 & 91.55 & 16505 & 8713 & 293 & 77.93 & 494 & 73.96 & 25.6 \\
    StrongSORT~\citep{18}    & \cmark & 72.38 & 85.22 & 16819 & 8559 & 256 & 80.55 & 188 & 73.88 & 21.3 \\
    ByteTrack~\citep{17}     & --     & 72.15 & 91.72 & 18055 & 7972 & 45 & 79.81 & 180 & 72.77 & 46.0 \\
    BoostTrack~\citep{64} & \cmark & 71.68 & 92.45 & 16611 & 8654 & 81 & 78.23 & 189 & \textbf{74.25} & 24.2 \\
    MOMT~\citep{49} & --     & 72.96 & \textbf{93.38} & 17319 & 8264 & 49 & \textbf{80.93} & 187 & 73.53 & 44.5 \\
    \textbf{DMSORT (ours)} & \cmark & \textbf{73.44} & 92.98 & \textbf{16479} & 8597 & 53 & 80.90 & 172 & 74.23 & 28.1 \\
    \hline
    % ---------------- 带星号（除 DMSORT* 外）全部红色 ----------------
    BOT-SORT$^\star$        & \cmark & 75.47 & 92.80 & 14161 & 11311 & 52 & 82.35 & \textbf{179} & 75.92 & 24.2 \\
    OCSORT$^\star$         & --     & 74.88 & 91.82 & 13551 & 10692 & 98 & 82.87 & 201 & 76.82 & \textbf{46.6} \\
    DeepOCSORT$^\star$     & \cmark & 75.89 & 91.96 & 12572 & 11321 & 100 & 82.96 & 195 & 77.61 & 29.6 \\
    ImprAssoc$^\star$      & \cmark & 75.53 & 91.92 & \textbf{12401} & 11753 & 202 & 82.06 & 377 & 77.51 & 28.3 \\
    StrongSORT$^\star$     & \cmark & 75.75 & 91.95 & 12549 & 11448 & 173 & 82.98 & 193 & 77.46 & 24.6 \\
    ByteTrack$^\star$      & --    & 75.24 & 91.68 & 12659 & 11430 & 66 & 83.50 & 199 & 77.53 & 45.3 \\
    BoostTrack$^\star$      & \cmark     & 75.20 & 92.11 & 13577 & 11388 & 91 & 83.11 & 191 & 76.89 & 27.2 \\
   MOMT$^\star$        & --     & 75.41 & 92.24 & 12468 & 11642 & 84 & 83.61 & 217 & \textbf{77.67} & 43.8 \\
    % ---------------- DMSORT* 保持黑色 ----------------
    \textbf{DMSORT$^\star$ (ours)} & \cmark & \textbf{76.69} & \textbf{93.25} & 13489 & \textbf{10425} & \textbf{40} & \textbf{83.67} & 196 & 77.60 & 30.6 \\
    \hline
  \end{tabular}%
  }
\end{table*}

\begin{figure*}
  \centering
  \subfloat{\includegraphics[width=0.95\columnwidth]{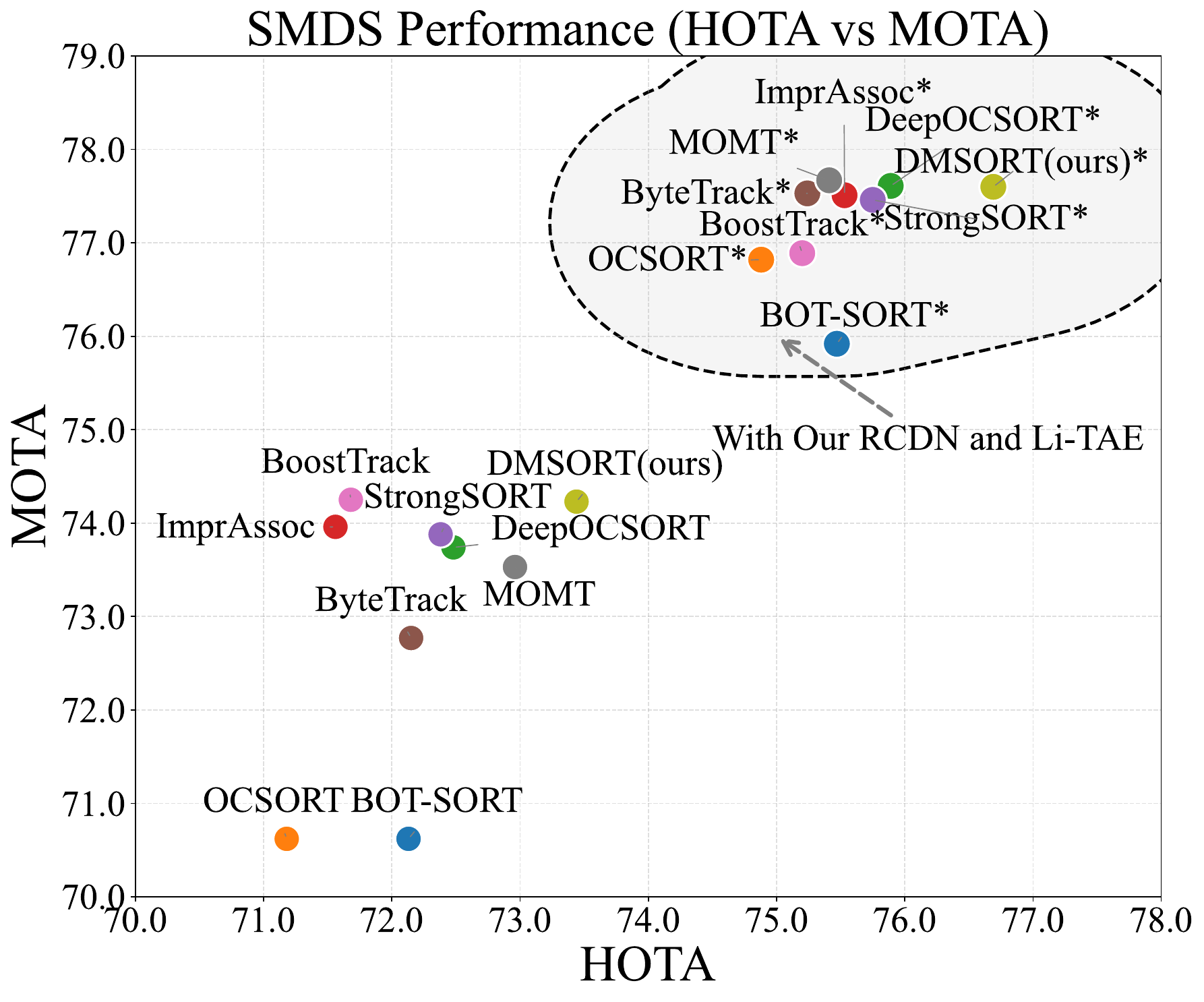}}\hspace{1mm}%
  \subfloat{\includegraphics[width=0.95\columnwidth]{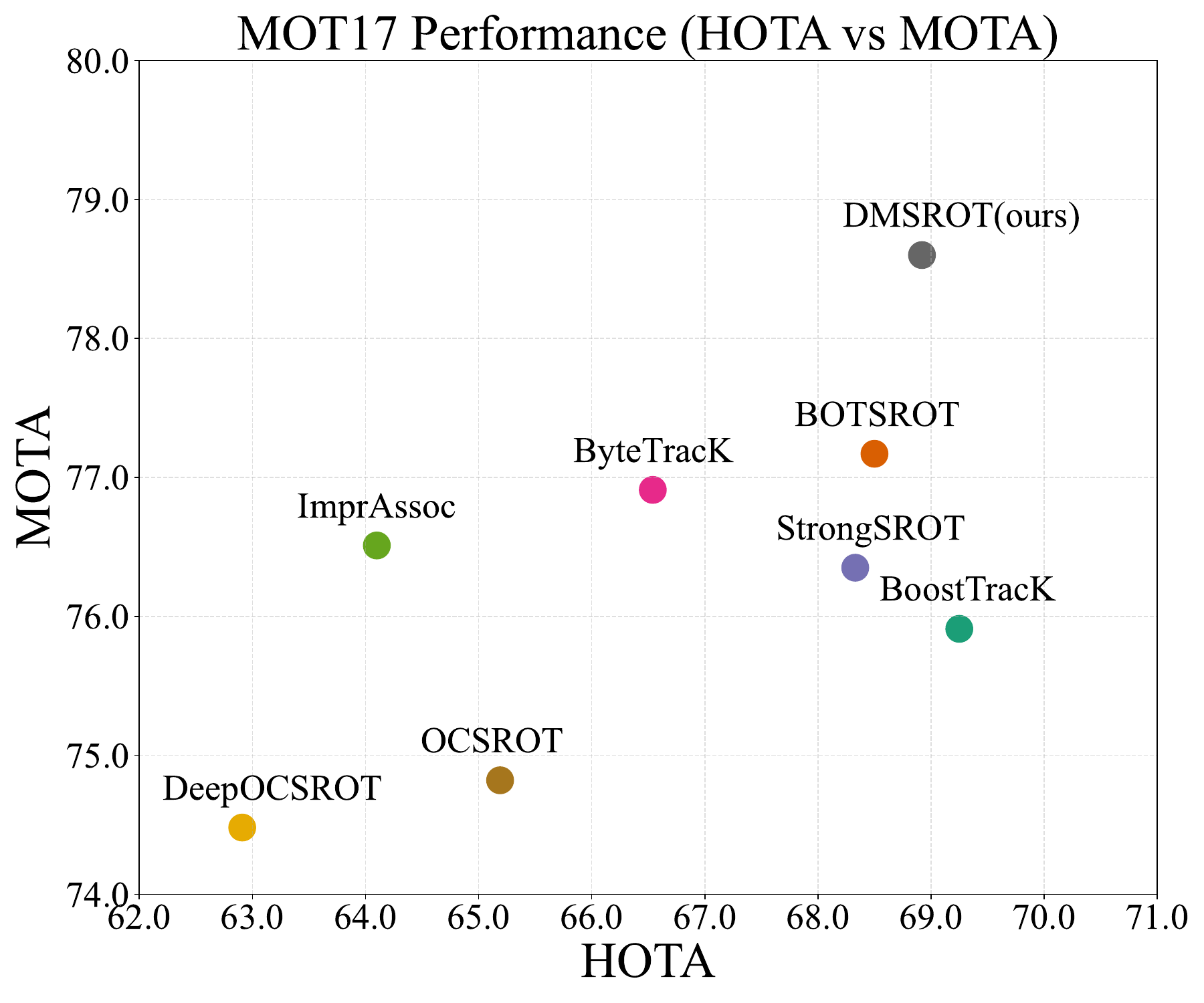}}
  \caption{Comparison of multi-object tracking algorithms on the Singapore Maritime Dataset (SMDS) and MOT17. Left: Results on SMDS using YOLOv11n with its default ReID model. Methods marked with an asterisk (*) are enhanced with RCDN and Li-TAE. Dashed ellipses group each method with its enhanced version for direct comparison; DMSORT denotes our complete method. Right: MOT17 results using detection and ReID models from BoxMOT without enhancement.}
  \label{fig:singapore_mot17_comparison}
\end{figure*}

In this study, we benchmark our proposed DMSORT against eight state-of-the-art trackers—BoT-SORT, OCSORT, DeepOCSORT, ImprAssoc, StrongSORT, ByteTrack, BoostTrack, and MOMT—using the Singapore Maritime Dataset (Table~\ref{tab:singapore_maritime} and Fig.~\ref{fig:singapore_mot17_comparison}). To ensure a fair comparison, all trackers use the same YOLOv11n detector. We further evaluate an enhanced version of each method (denoted with $\star$) that integrates our proposed RCDN backbone and, where applicable, the Li-TAE appearance encoder.

It is noteworthy that all $\star$-enhanced methods exhibit noticeable improvements across multiple tracking metrics compared to their vanilla counterparts, demonstrating the general effectiveness and plug-and-play capability of both the RCDN backbone and the Li-TAE module. For instance, DeepOCSORT$^\star$ achieves a 3.41\% increase in HOTA and 3.87\% in MOTA, while BOT-SORT$^\star$ improves IDF1 by 3.54\% and reduces false negatives significantly. These consistent gains across different tracking paradigms highlight the broad applicability of our proposed components.

Our base DMSORT already demonstrates competitive performance, achieving the highest HOTA (73.44\%) and lowest false negatives (16,479) among the original methods. More importantly, the enhanced DMSORT$^\star$ consistently outperforms all $\star$-enhanced counterparts across most key metrics. It attains the best HOTA (76.69\%), IDF1 (83.67\%), and localization accuracy (LocA = 93.25\%), while also yielding the fewest false negatives (13,489) and fragments (40). Although its MOTA (77.60\%) is slightly lower than the top performer MOMT$^\star$ (77.67\%), the difference is marginal at only 0.07\%.

Regarding localization accuracy (LocA), while RCDN is theoretically designed to enhance localization through its reversible architecture and feature preservation, some $\star$-enhanced methods show slight decreases in LocA. This can be attributed to the increased detection sensitivity brought by RCDN, which may introduce more challenging localization cases (e.g., partially occluded or truncated objects) that slightly affect average precision. However, our DMSORT$^\star$ still achieves the highest LocA (93.25\%), demonstrating its superior localization capability.

In terms of inference speed, while RCDN has smaller parameter counts and FLOPs than the baseline detector, its deeper architecture may result in slightly slower FPS compared to YOLOv11. However, the lightweight Li-TAE module significantly improves the inference speed for all Re-ID-based methods, as evidenced by the FPS improvements in most $\star$-enhanced trackers.

For identity switches (IDs) and fragments (Frag), the $\star$-enhanced versions generally exhibit increases in these metrics, which can be attributed to the substantially improved detection sensitivity of RCDN. While the higher recall successfully reduces false negatives (FN), it also introduces more detected objects—including those that are partially occluded, short-lived, or in close proximity—which inherently increases the complexity of data association. This, in turn, may lead to more frequent track fragmentations and identity switches. Nevertheless, this represents a favorable trade-off: the significant enhancement in recall and the reduction in missed detections far outweigh the moderate increase in IDs and Frag, ultimately contributing to more robust and continuous tracking in challenging maritime environments.

Overall, by combining high-recall detection, the compact yet powerful Li-TAE appearance encoder, and our proposed COFF module, DMSORT$^\star$ achieves balanced and state-of-the-art performance in cluttered maritime scenes—excelling in detection, association, localization, and temporal stability.

\begin{table}
  \centering
  \caption{Comparison Experiment of DMSORT on the MOT17 Dataset (Best in Bold)}
  \label{tab:mot17}
  \setlength{\tabcolsep}{8pt}   % 与原表相同
  \renewcommand{\arraystretch}{1.3}% 与原表相同
  \resizebox{0.9\columnwidth}{!}{%
  \begin{tabular}{@{}l|c|c|c|c@{}}
    \hline
    Tracker & Re-ID & HOTA$\uparrow$ & MOTA$\uparrow$ & IDF1$\uparrow$ \\
    \hline
    BoostTrack~\citep{64} & \cmark & \textbf{69.25} & 75.91 & \textbf{83.21} \\
    BOT-SORT~\citep{21}   & \cmark & 68.50 & 77.17 & 80.99 \\
    StrongSORT~\citep{18} & \cmark & 68.33 & 76.35 & 81.21 \\
    ByteTrack~\citep{17}  & -- & 66.54 & 76.91 & 77.86 \\
    ImprAssoc~\citep{44}  & \cmark & 64.10 & 76.51 & 71.88 \\
    DeepOCSORT~\citep{45} & \cmark & 62.91 & 74.48 & 73.46 \\
    OCSORT~\citep{43}     & -- & 65.19 & 74.82 & 75.96 \\
    DMSORT (ours) & \cmark & 68.92 & \textbf{78.60} & 81.57 \\
    \hline
  \end{tabular}%
  }
\end{table}

To further assess the generalization capability of our maritime-specific tracker, we evaluate DMSORT on the widely adopted MOT17 benchmark. In line with BOXMOT, we report the three most critical tracking metrics—MOTA, HOTA, and IDF1—as shown in Table~\ref{tab:mot17} and Fig.~\ref{fig:singapore_mot17_comparison}. Despite being tailored for maritime multi-object tracking, DMSORT demonstrates strong generalization to crowded, occlusion-heavy urban scenes. Specifically, it achieves the highest MOTA score (78.60), outperforming all other state-of-the-art methods, including those explicitly optimized for MOT17. Additionally, it ranks second in both HOTA (68.92) and IDF1 (81.57), closely trailing BoostTrack, which attains marginally higher values.

These results confirm that our modular design comprising high-recall detection, appearance modeling via Li-TAE, and robust motion association with COFF remains effective even under drastically different visual domains. The consistent performance across maritime and pedestrian tracking scenarios highlights DMSORT’s strong adaptability and robustness in multi-object tracking tasks.

\subsection{ Effectiveness Analysis of COFF}
\begin{figure}
    \centering
    \includegraphics[width=0.9\columnwidth]{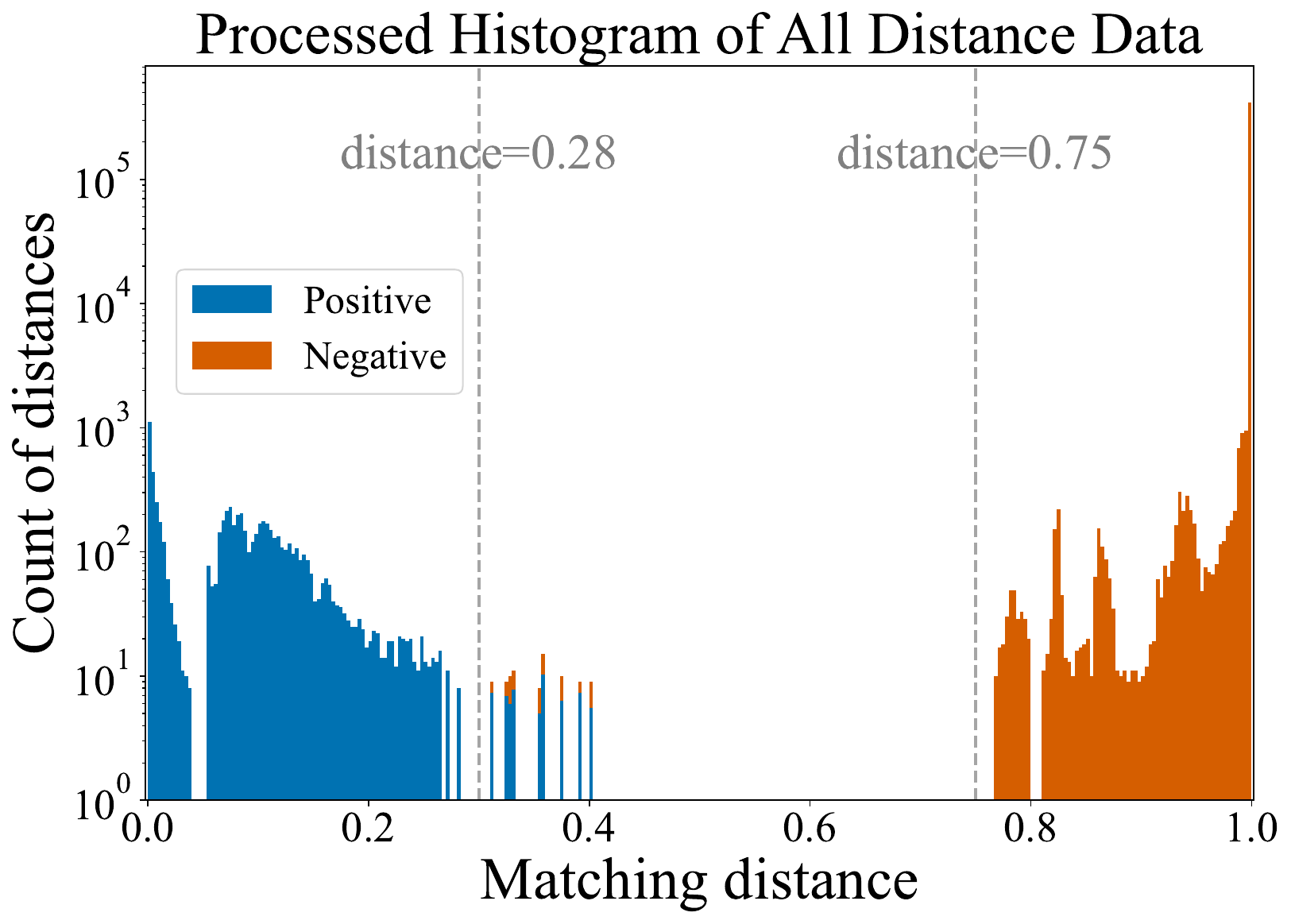}
    \caption{Distribution of positive and negative samples post-feature clustering on the Singapore Maritime Dataset. Positive samples (blue) cluster closely below 0.28, indicating high identity similarity. Negative samples (orange) predominantly gather near 1, showing clear separation. This distribution underscores the effectiveness of feature clustering in enhancing tracking accuracy and robustness.}
    \label{fig:metrics_jvlei}
\end{figure}

Fig.~\ref{fig:metrics_jvlei} illustrates the distribution of positive and negative samples during the matching process on the Singapore Maritime Dataset.The blue bars represent positive samples (matching the same identity), while the orange bars denote negative samples (non-matching identities). The x-axis represents the distance between samples, ranging from 0 to 1, while the y-axis shows the processed frequency on a logarithmic scale. The distances are divided into 300 intervals, with bins containing fewer than 8 samples excluded to minimize outlier effects.

The plot reveals that after applying feature clustering, negative samples predominantly cluster around a distance of 1, with over 95\% falling within the [0.95, 1.0] range, indicating effective separation from the positive samples. In contrast, positive samples are tightly clustered within the [0, 0.28] range, reflecting high feature similarity among the same identity. The intermediate range (0.28, 0.75) contains sparse occurrences, with fewer than 8 samples per bin, suggesting minimal confusion between positive and negative samples. This clear separation between the two classes demonstrates that feature clustering significantly enhances the discriminative power of the tracker, improving both accuracy and robustness in multi-object tracking.

\begin{figure*}
  \centering
  \subfloat{\includegraphics[width=0.95\columnwidth]{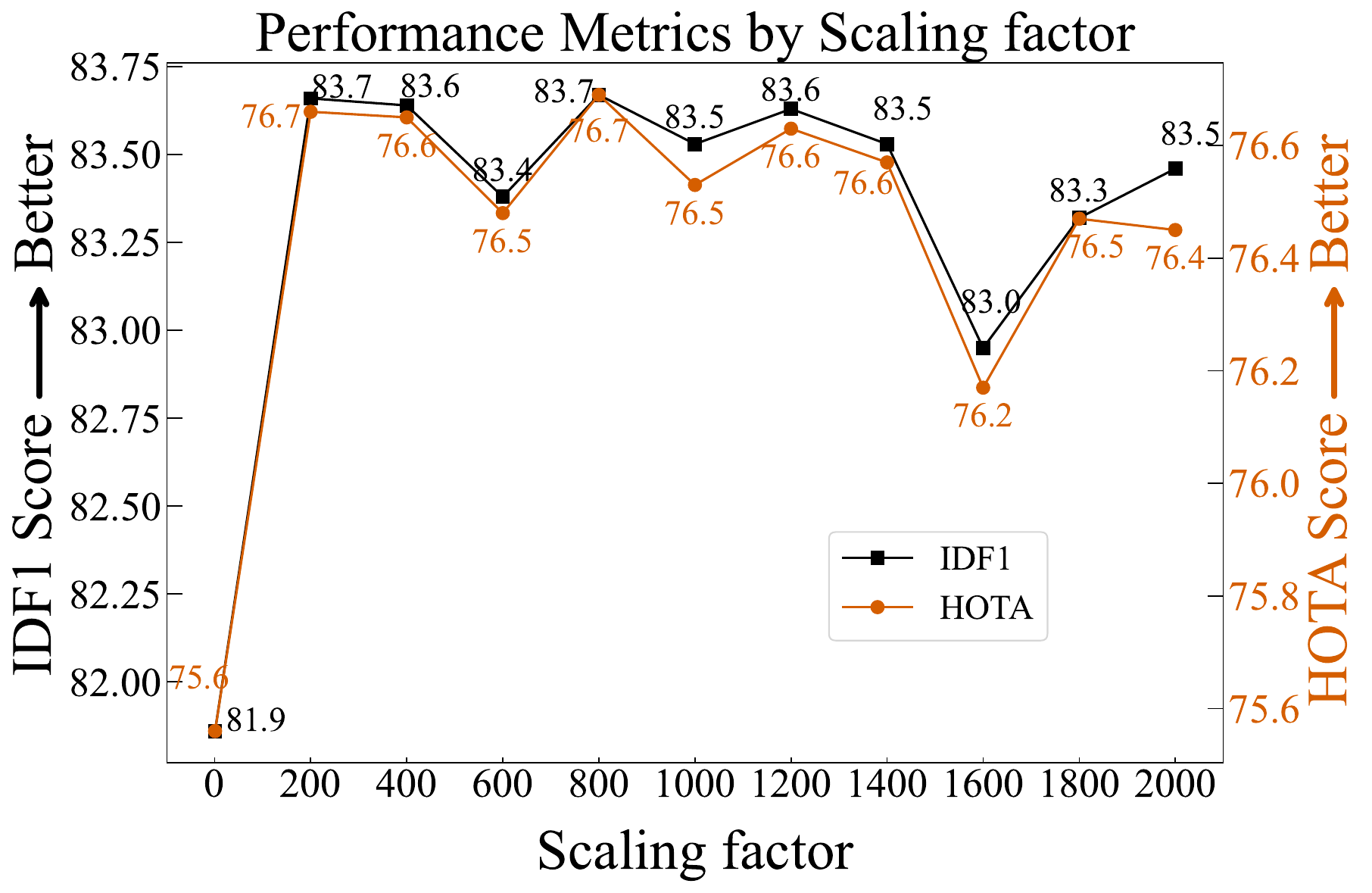}}\hspace{1mm}%
  \subfloat{\includegraphics[width=0.95\columnwidth]{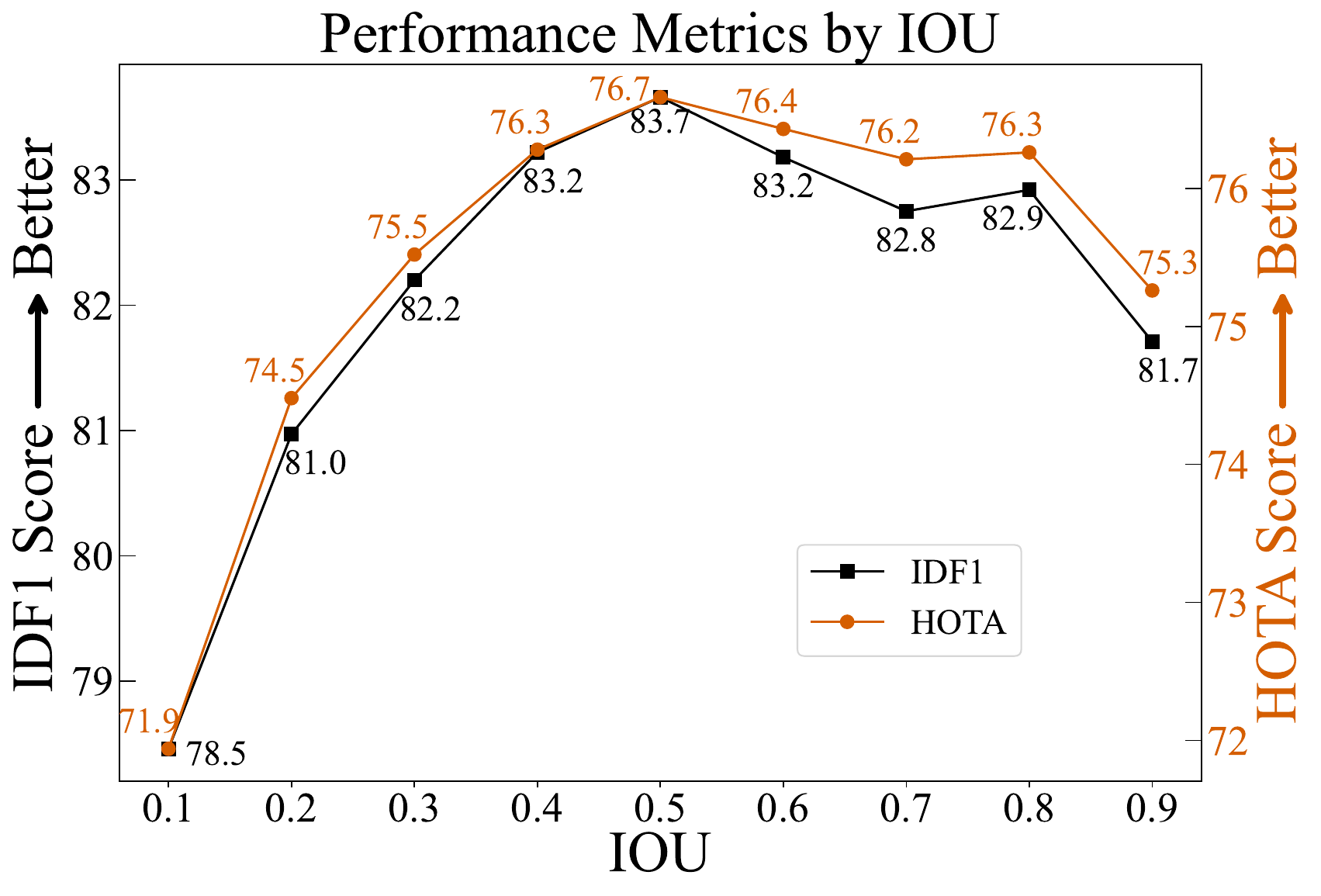}}
  \caption{Evaluation of the DMSORT tracking method using IDF1 and HOTA metrics: left at different scaling factor thresholds during the matching phase, and right at different IOU thresholds during the detector inference phase with non-maximum suppression.}
  \label{fig:dis_jvlei_iou}
\end{figure*}

Fig.~\ref{fig:dis_jvlei_iou} presents IDF1 and HOTA as the amplification factor $\beta$ increases from 200 to 2000 (a ten-fold span). Across this range, both metrics remain remarkably stable and consistently exceed the “no feature-scaling” baseline (HOTA = 75.6\%, IDF1 = 82.8\%). Specifically, HOTA varies only between 76.2\% and 76.7\%—an absolute gain of approximately 1\% ,while IDF1 ranges from 83.0\% to 83.7\%, peaking at $\beta$ = 800. Relative to the unscaled features, our clustering-optimized fusion yields a consistent improvement of about 1.0\% in HOTA and 0.9\% in IDF1, demonstrating that the proposed scaling is both low-sensitivity and reliably beneficial.

\subsection{ Detector Analysis}

\begin{table*}
  \centering
  \caption{Comparison of Model Performance in Detection and Tracking}
  \label{tab:combined_model_comparison}
  \setlength{\tabcolsep}{4.5pt}   % 与原表相同
  \renewcommand{\arraystretch}{1.2}% 与原表相同
  \resizebox{0.9\textwidth}{!}{%
  \begin{tabular}{@{}l|cc|cccc|ccc@{}}
    \hline
    \multirow{2}{*}{Model} & \multicolumn{2}{c|}{Parameters and FLOPs} & \multicolumn{4}{c|}{Detection Performance} & \multicolumn{3}{c}{Tracking Performance} \\
    \cline{2-10}
    & Params & FLOPs & Precision & Recall Rate & mAP 50\% & mAP 50-95\% & MOTA$\uparrow$ & IDF1$\uparrow$ & HOTA \\
    \hline
    RCDN (ours) & 2.2\,M & 5.0\,G & 0.9316 & 0.8812 & 0.9332 & 0.7305 & 77.60 & 83.67 & 76.69 \\
    YOLOv11-CAF &
3.5\,M &10.9\,G &
0.9369 &0.9025 &
0.9425 &0.6900 &
76.24 &83.41 &
76.68 \\
    YOLOv11-tiny & 2.6\,M & 6.5\,G & 0.9264 & 0.8503 & 0.9116 & 0.6771 & 75.68 & 82.00 & 74.42 \\
    YOLOv11-small & 9.4\,M & 21.6\,G & 0.9570 & 0.9165 & 0.9633 & 0.7698 & 79.38 & 85.19 & 79.38 \\
    YOLOv11-medium & 20.1\,M & 68.4\,G & 0.9692 & 0.9305 & 0.9706 & 0.7608 & 79.91 & 85.82 & 79.20 \\
    \hline
  \end{tabular}%
  }
\end{table*}

Under the Detection-Before-Tracking (DBT) paradigm, the fidelity of the object detector directly influences the downstream multi-object tracking (MOT) performance. As shown in Table~\ref{tab:combined_model_comparison}, scaling the YOLOv11 family from tiny (6.5G FLOPs) to small (21.6G FLOPs) leads to substantial improvements: MOTA increases by 3.7\% , IDF1 by 3.19\% , and HOTA by 4.96\%. These consistent gains highlight the benefits of increased detector capacity in enhancing feature quality, which translates into better tracking completeness and identity preservation in complex maritime scenes.

However, further scaling from YOLOv11-small (21.6G FLOPs) to YOLOv11-medium (68.4G FLOPs) yields only marginal improvements: MOTA rises by just 0.53\%, IDF1 by 0.63\%, and HOTA shows negligible change (from 79.38 to 79.20). This diminishing return suggests that brute-force model enlargement quickly reaches a point of limited utility and highlights the need for architectural innovations that deliver stronger accuracy-efficiency trade-offs.

Although YOLOv11-CAF~\cite{67} and our RCDN share the goal of enhancing information flow and preserving features, RCDN delivers superior multi-object tracking performance. Its reversible architecture ensures lossless feature transformation and yields higher mAP (0.7305 vs 0.6900) while requiring only 5.0 G FLOPs—less than half of YOLOv11-CAF’s 10.9 G—making it better suited for resource-constrained maritime scenarios.

To this end, we propose the Reversible Columnar Detection Network (RCDN), which, with only 5.0G FLOPs, achieves a HOTA of 76.69—just 2.69\% below YOLOv11-small yet with less than one-quarter of its computational cost. Compared to YOLOv11-tiny, RCDN improves HOTA by 2.27\% while using fewer FLOPs. These results demonstrate that RCDN effectively bridges the gap between compact and mid-sized detectors. Its reversible columnar architecture and cross-scale feature reuse mechanism enable efficient bi-directional flow and fusion of multi-resolution features, capturing rich spatial semantics with minimal overhead. This lightweight yet performant design is well suited for maritime platforms with constrained resources, such as embedded vessel trackers and autonomous buoys, where both computational efficiency and robust multi-object tracking are essential.

Fig.~\ref{fig:dis_jvlei_iou} presents a sensitivity analysis of the model’s performance under varying Intersection-over-Union (IOU) thresholds, illustrating non-monotonic trends in both IDF1 and HOTA scores. At low IOU values (0.1–0.3), IDF1 ranges from 78.5\% to 82.2\% and HOTA from 71.9\% to 75.5\%, indicating that overly lenient matching allows more false positives, which degrades target discrimination and tracking accuracy. As the IOU threshold increases into the moderate range (0.4–0.6), both metrics improve markedly—IDF1 peaks at 83.7\% and HOTA at 76.7\%—showing that this range strikes an optimal balance between acceptance of true matches and rejection of false ones. Beyond 0.7, overly strict criteria begin to exclude valid associations, causing IDF1 to fall back to 81.7\%–82.9\% and HOTA to 75.3\%–76.3\%.

Taken together, these results highlight that an IOU threshold of 0.5 yields the best overall performance, achieving the highest IDF1 and HOTA scores while maintaining robust target discrimination and reliable tracking. This underscores the critical importance of selecting an appropriate IOU threshold to optimize multi-object tracking systems.

\begin{figure*}
  \centering

  \includegraphics[width=0.9\textwidth]{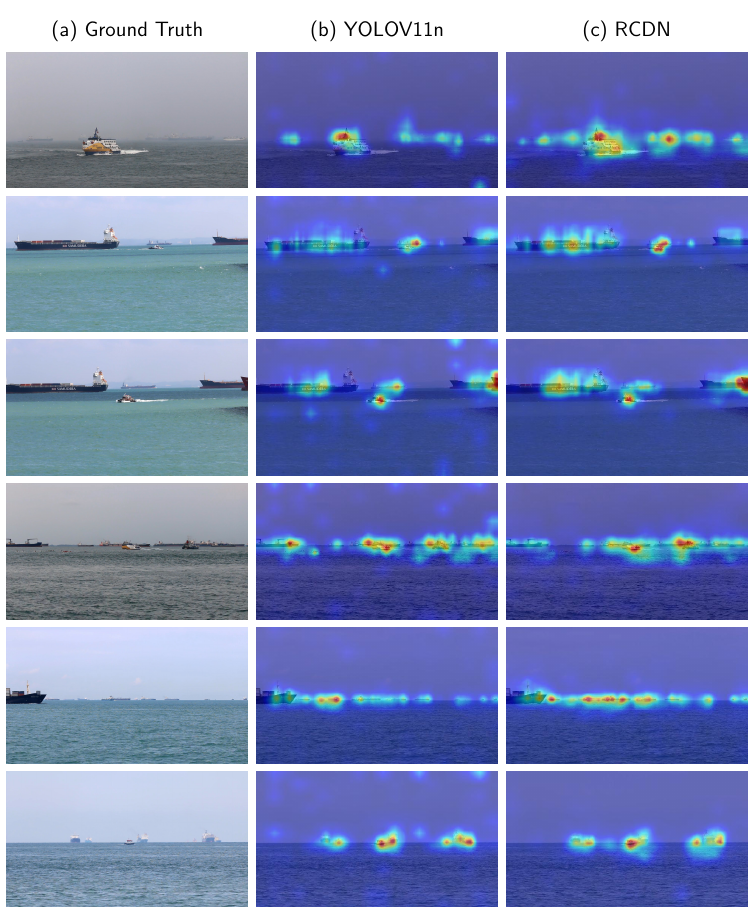}

  \caption{Comparison of detection heatmaps under various maritime conditions using the proposed Reversible Columnar Detection Network (RCDN) and YOLOv11n. Scenarios include normal weather, fog, and dense small targets.}
  \label{fig:multi_image_grid}
\end{figure*}

Through the visualization of layer heatmaps Fig.~\ref{fig:multi_image_grid}, our proposed RCDN demonstrates clear superiority in dense small-object detection, blurred scenarios, and standard maritime environments. Specifically, RCDN extends its receptive focus across multiple closely spaced targets—outperforming YOLOv11 in robustness and localization accuracy when vessels congregate—while simultaneously exhibiting heightened sensitivity to diminutive objects, such as small boats and fishing vessels, that are easily overlooked in low-contrast or noisy backgrounds. Moreover, RCDN substantially suppresses attention to irrelevant regions (e.g., sea surface and sky), thereby mitigating environmental noise effects and producing sharply concentrated attention zones even amid complex backgrounds like waves and cloud cover. These characteristics underscore RCDN’s enhanced precision and resilience in challenging maritime detection tasks.

\subsection{Visualization Results’ Analysis}

\begin{figure*}          % 与原 htbp 等价
  \centering

  \includegraphics[width=0.9\textwidth]{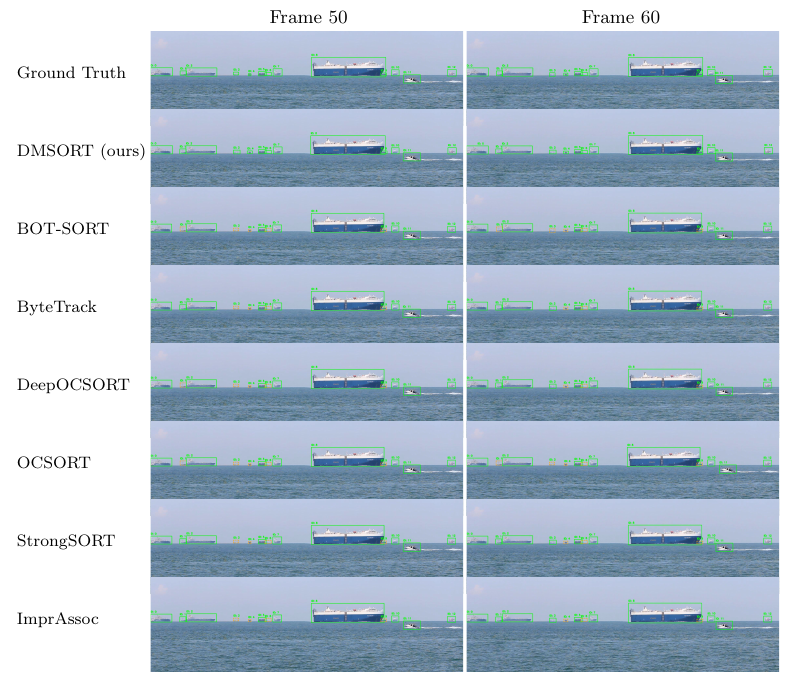}
  \vspace{1mm}

  \caption{Visual comparison of different detection and tracking algorithms with Kalman filter updates in a bright environment featuring many objects of varying sizes intertwined. Yellow dashed boxes indicate lost tracking objects.}
  \label{fig:track_hard}
\end{figure*}

To further evaluate the robustness of our proposed framework in realistic maritime scenarios, we present a visual comparison of tracking performance across sequential frames in challenging scenes characterized by dense small targets and significant motion blur. As illustrated in Fig.~\ref{fig:track_hard}, most existing trackers suffer from delayed or incomplete trajectory initialization—many small and visually degraded objects remain undetected or untracked until after frame 60, and some fail to form valid trajectories throughout the entire sequence. This reflects the limitations of their detection sensitivity and long-term association stability under severe maritime degradation. In contrast, our DMSORT successfully establishes stable and continuous trajectories as early as frame 50, including for low-resolution, occluded targets. This improvement stems from the synergy between two key innovations: the Reversible Columnar Detection Network (RCDN), which ensures high recall through progressive multi-scale feature refinement, and the clustering optimized feature fusion (COFF) strategy, which enables resilient data association by adaptively balancing motion and appearance cues through nonlinear fusion and spatial gating. Together, these components allow DMSORT to maintain identity consistency, accelerate trajectory formation, and reduce fragmentation in highly dynamic maritime conditions. This advantage is particularly evident in early frames where competing methods either delay tracking or lose targets altogether, confirming the efficacy of our detection-association design in supporting persistent maritime multi-object tracking.

\begin{figure*}
  \centering
  \includegraphics[width=0.9\textwidth]{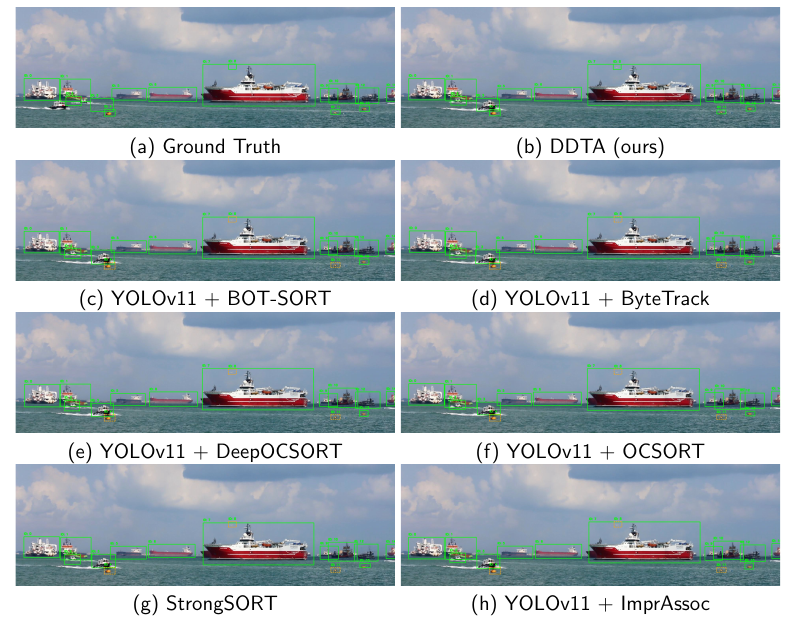}

  \caption{Visual comparison of different detection and tracking algorithms with Kalman filter updates in a bright environment featuring many objects of varying sizes intertwined. Yellow dashed boxes indicate lost tracking objects.}
  \label{fig:dection}
\end{figure*}
The Fig.~\ref{fig:dection} provides a visual comparison of different detection and tracking algorithms using Kalman filter updates in a bright maritime environment. From this, we can clearly observe that our method outperforms the others in terms of detection, particularly in higher recall rates and more accurate bounding boxes. Specifically, the RCDN's reversible architecture and precise feature representation enable better detection of small maritime targets, significantly boosting the recall rate. Additionally, our method benefits from robust Kalman filtering, which is facilitated by accurate camera motion estimation and stable detection. The camera estimation effectively reduces errors in the Kalman filter update process, leading to more reliable tracking and minimizing trajectory fragmentation. As a result, the Kalman filter becomes more stable, reducing accumulated uncertainty and improving detection accuracy. In contrast, other methods suffer from issues such as trajectory fragmentation and incorrect associations, which destabilize the Kalman filter process, introducing noise that disrupts target motion estimation.

\begin{figure*}
    \centering
    \includegraphics[width=0.9\textwidth]{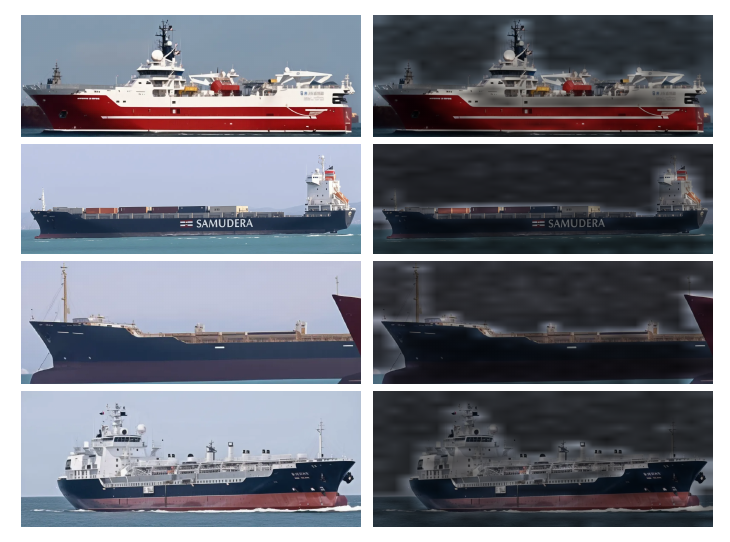}

    \caption{Visualization of the LiTAE attention map on sample frames from the Singapore Maritime Dataset. For each pair of images, the left one is the original frame, and the right one is the corresponding attention - mapped frame. In the attention - mapped frames, regions with larger attention weights are relatively brighter, while regions with smaller attention weights are darker.}
    \label{fig:li_tae_vis}
\end{figure*}

The Fig.~\ref{fig:li_tae_vis} illustrates the attention map produced by the Li-TAE model on ship images extracted from the Singapore Maritime Dataset. These attention maps are generated using multi-head averaging to highlight the focus of the model on key vessel features.After the ships are detected and cropped, they are fed into the Li-TAE model to extract their re-identification features. The attention map, generated using a multi-head attention strategy, reveals how the model allocates its focus on different regions of the ship. As shown in the figure, our Li-TAE model, after being trained with adaptive hard triplets, effectively captures the vessel's key features with near segmentation-level accuracy. This capability is a result of the global receptive field facilitated by self-attention mechanisms, which works exceptionally well in maritime environments. The model demonstrates robust feature extraction, ensuring excellent performance even under challenging oceanic conditions.

\subsection{Ablation Study}

\begin{table*}
  \centering
  \caption{Ablation Study of DMSORT on the Singapore Maritime Dataset}
  \label{tab:ablation_study}
  {\fontsize{8pt}{9pt}\selectfont
  \setlength{\tabcolsep}{4pt}
  \renewcommand{\arraystretch}{1.1}
  \resizebox{0.85\textwidth}{!}{%
    \begin{tabular}{l|cccc|cccc}
      \hline
       Method & Li-TAE & DDTA & COFF & RCDN & HOTA\rlap{↑} & MOTA\rlap{↑} & IDF1\rlap{↑} & FPS\rlap{↑} \\
      \hline
      Baseline (BOT-SORT) & -- & -- & -- & -- & 72.13 & 70.62 & 78.81 & 20.4 \\
      Baseline + col. 1 & \checkmark & -- & -- & -- & 72.96 & 74.00 & 80.67 & 24.6 \\
      Baseline + col. 1--2 & \checkmark & \checkmark & -- & -- & 72.98 & 74.00& 80.68 & \textbf{31.2} \\
      Baseline + col. 1--3 & \checkmark & \checkmark & \checkmark & -- & 74.42 & 75.68 & 82.00 & 31.1 \\
      Baseline + col. 1--4 & \checkmark & \checkmark & \checkmark & \checkmark & \textbf{76.69} & \textbf{77.60} & \textbf{83.67} & 30.6 \\
      \hline
    \end{tabular}%
  }
}
\end{table*}

As shown in Table \ref{tab:ablation_study}, introducing Li-TAE immediately enhances feature discrimination—MOTA jumps by 3.38\% and IDF1 by 1.86\% thanks to its global receptive field and adaptive angular margin loss. Adding DDTA further stabilizes long-term track continuity under affine transformations while boosting throughput to 31.2 FPS. With COFF, the fusion of clustered motion cues and appearance embeddings yields an extra +1.44\% in HOTA and +1.32\% in IDF1, reducing identity switches in crowded scenes. Finally, RCDN’s reversible, columnar detection–tracking co-design elevates detection reliability, driving HOTA up to 76.69\% and IDF1 to 83.67\%. This incremental ablation confirms that each innovation—appearance encoding, affine-aware tracking, motion–appearance fusion, and joint optimization—contributes uniquely to our system’s robust, real-time MOT performance.

\section{Conclusion}
This paper presents DMSORT, a domain-adapted multi-object tracking framework specifically designed for the challenges of maritime environments, including severe occlusion, target scale variation, camera motion, and visual degradation. The proposed system integrates four tightly coupled innovations: a Dual-Branch Detection-Tracking Architecture (DDTA) for real-time motion-aware tracking, a Reversible Columnar Detection Network (RCDN) for multi-scale feature refinement, the Li-TAE Transformer-based appearance encoder for robust identity modeling, and a Clustering Optimized Feature Fusion (COFF) strategy for resilient association under uncertainty.

Through extensive evaluation on the Singapore Maritime Dataset, DMSORT sets a new state-of-the-art with 76.69 HOTA, 83.67 IDF1, and 77.60 MOTA, outperforming existing SOTA trackers in identity consistency, detection quality, and trajectory stability. Notably, DMSORT also achieves strong cross-domain generalization on MOT17, surpassing urban-focused methods in MOTA (78.60) and performing competitively in HOTA and IDF1, despite its maritime specialization.

These results validate our key design principles: decoupling detection and tracking for parallelism and motion compensation, enhancing detection with reversible multi-scale backbones, leveraging global attention for discriminative Re-ID, and fusing cues through a learned, nonlinear association space. Together, these modules form a modular and efficient tracking solution with broad applicability, from maritime surveillance to general-purpose MOT.

In future work, we plan to explore scalable multi-camera deployment, incorporate temporal modeling for long-term occlusion handling, and extend the framework toward radar-visual multimodal fusion. This integration aims to leverage radar's robustness in adverse conditions and long-range perception, complementing RGB cues for reliable all-weather maritime tracking.

%%%%%%%%% REFERENCES
{
% \fontsize{8.2pt}{9.84pt}\selectfont
% \bibliographystyle{unsrt}
\bibliography{egbib}
\bibliographystyle{cas-model2-names}  % 或者 apalike、plainna
% \bibliography{regbib}
}

\end{document}